%% file: arxiv.tex
\documentclass[runningheads]{llncs}

\usepackage{eccv}

\usepackage{eccvabbrv}

\usepackage{graphicx}
\usepackage{booktabs}

\usepackage[accsupp]{axessibility}  

\usepackage[pagebackref,breaklinks,colorlinks,citecolor=eccvblue]{hyperref}

\usepackage{orcidlink}

\input{preamble}

\begin{document}

\title{Mesh2NeRF: Direct Mesh Supervision for Neural Radiance Field Representation and Generation} 

\titlerunning{Mesh2NeRF}

\author{Yujin Chen\inst{1} \and Yinyu Nie\inst{1} \and Benjamin Ummenhofer\inst{2} \and Reiner Birkl\inst{2} \and Michael~Paulitsch\inst{2} \and Matthias Müller\inst{2} \and
Matthias Nießner\inst{1}}

\authorrunning{Y.Chen et al.}
\institute{Technical University of Munich \and
Intel Labs}

\maketitle
\vspace{-0.1in}
\begin{center}
\includegraphics[width=1\textwidth]{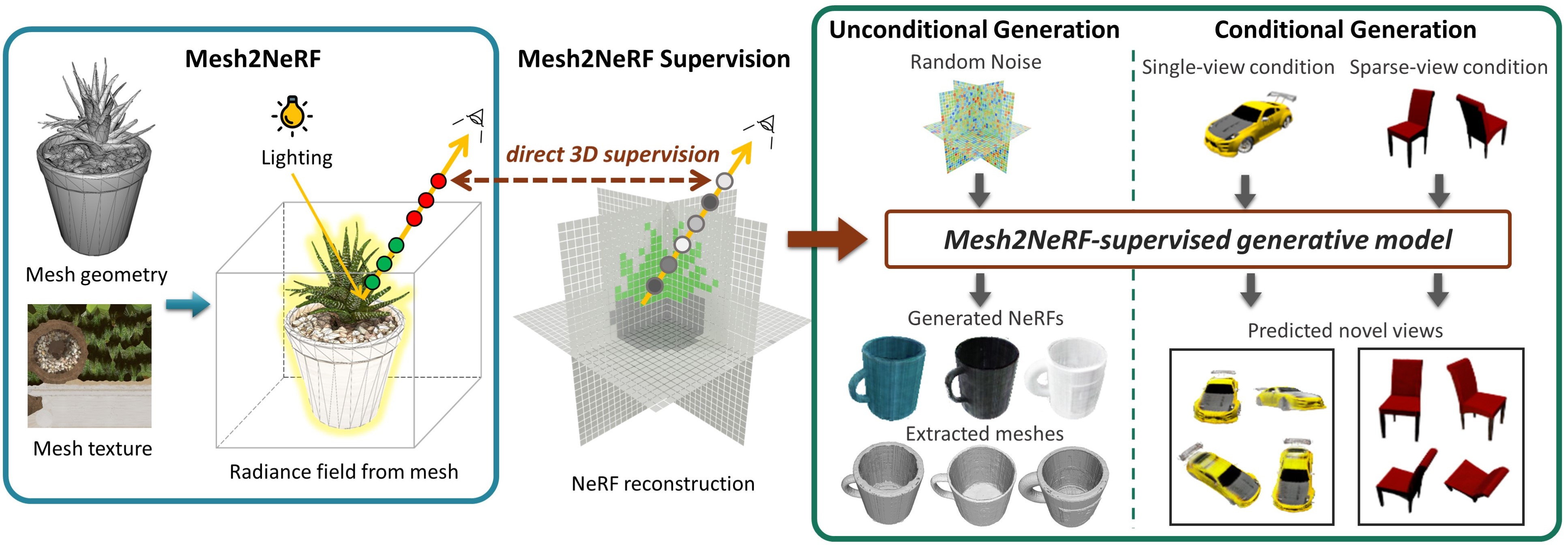}
\vspace{-0.2in}
\captionof{figure}{
We propose \OURS{}, a novel method for extracting ground truth radiance fields directly from 3D textured meshes by incorporating mesh geometry, texture, and environment lighting information. 
\OURS serves as direct 3D supervision for neural radiance fields, offering a comprehensive approach to leveraging mesh data for improving novel view synthesis performance.
\OURS{} can function as supervision for generative models during training on mesh collections, advancing various 3D generation tasks, including unconditional and conditional generation.
}
\label{fig:pipeline}
\end{center}

\input{AuthorKit/sections_eccv/0abstract_eccv}
\input{AuthorKit/sections_eccv/1introduction_eccv}
\input{AuthorKit/sections_eccv/2relatedworks_eccv}
\input{AuthorKit/sections_eccv/3method_eccv}

\input{AuthorKit/sections_eccv/4results_eccv}
\input{AuthorKit/sections_eccv/5conclusion_eccv}

\noindent \textbf{Acknowledgements.} This work is mainly supported by a gift from Intel. It was also supported by the ERC Starting Grant Scan2CAD (804724) as well as the German Research Foundation (DFG) Research Unit ``Learning and Simulation in Visual Computing''.

\bibliographystyle{splncs04}
\bibliography{main}

\input{AuthorKit/sections_eccv/6appendix_eccv}

\end{document}

%% file: preamble.tex
\usepackage{epsfig}
\usepackage{graphicx}
\usepackage{amsmath}
\usepackage{amssymb}
\usepackage{multirow}
\usepackage{booktabs}
\usepackage{comment}
\usepackage{subcaption}
\usepackage{makecell}
\usepackage{wrapfig}

\newcommand\mypara[1]{\vspace{1mm}\noindent\textbf{#1.}}

\newcommand{\OURS}{Mesh2NeRF\xspace}
\definecolor{darkgreen}{RGB}{65,162,52}
\definecolor{darkblue}{RGB}{66,154,214}
\definecolor{lightgray}{RGB}{211,211,211}
\newcommand*{\aboverulesepcolor}[1]{%
  \noalign{%
    \begingroup 
      \color{#1}%
      \hrule height\aboverulesep 
    \endgroup 
    \kern-\aboverulesep 
  }%
} 

\definecolor{yinyu}{rgb}{0.6,0.3,0.9}

%% file: AuthorKit/sections_eccv/0abstract_eccv.tex
\begin{abstract}
We present \OURS, an approach to derive ground-truth radiance fields from textured meshes for 3D generation tasks. Many 3D generative approaches represent 3D scenes as radiance fields for training. Their ground-truth radiance fields are usually fitted from multi-view renderings from a large-scale synthetic 3D dataset, which often results in artifacts due to occlusions or under-fitting issues. In \OURS, we propose an analytic solution to directly obtain ground-truth radiance fields from 3D meshes, characterizing the density field with an occupancy function featuring a defined surface thickness, and determining view-dependent color through a reflection function considering both the mesh and environment lighting. \OURS extracts accurate radiance fields which provides direct supervision for training generative NeRFs and single scene representation. 
We validate the effectiveness of Mesh2NeRF across various tasks, achieving a noteworthy 3.12dB improvement in PSNR for view synthesis in single scene representation on the ABO dataset, a 0.69 PSNR enhancement in the single-view conditional generation of ShapeNet Cars, and notably improved mesh extraction from NeRF in the unconditional generation of Objaverse Mugs.

\keywords{Radiance Field Supervision \and NeRF Generation \and Mesh Prior}
\end{abstract}

%% file: AuthorKit/sections_eccv/1introduction_eccv.tex
\section{Introduction}
3D virtual content generation has garnered increased attention within the domains of computer vision and graphics. This surge is fueled by advancements in large-scale datasets~\cite{chang2015shapenet, deitke2023objaverse, dai2017scannet, collins2022abo, deitke2024objaverse} and generative models~\cite{chan2021pi, schwarz2020graf, chan2022efficient, niemeyer2021giraffe, rombach2022high}, fostering significant growth in the 3D asset creation industry. 
Classical 3D representations, including point clouds, meshes, voxel grids, SDF, etc., have seamlessly integrated into generative models, yielding promising results in generating 3D shapes~\cite{wu2016learning, kato2018neural, liu2019soft, sitzmann2019scene, niemeyer2020differentiable}.
Radiance fields have emerged as a powerful learning representation for 3D reconstruction and generation, as evidenced by their effectiveness in various studies~\cite{mildenhall2021nerf, wang2021neus, jang2021codenerf, sun2021neuralrecon, azinovic2022neural, lin2023vision, muller2023diffrf, ssdnerf}.

Radiance fields possess significant potential as high-quality 3D generative representations, but ground truth radiance fields are required as training samples for generative models.
However, obtaining GT radiance fields is extremely challenging. To overcome the lack of direct 3D radiance field supervision, some recent approaches utilize 2D generative models~\cite{nichol2021glide, rombach2022high, zhang2023adding, betker2023improving} to generate multi-view posed images~\cite{szymanowicz2023viewset, hoellein2024viewdiff}. Subsequently, these 2D generations are fitted into NeRFs or 4D grids (RGB + opacity) to obtain textured 3D outputs~\cite{liu2023zero, liu2023one2345++}. Nevertheless, these methods rely on the quality and consistency of upstream 2D multi-view generation, posing challenges in ensuring high-quality 3D outputs.

Some methods directly train a 3D generative model on native 3D data ~\cite{luo2021diffusion, shue20233d, chou2023diffusion, nam20223d}, leveraging inherent 3D supervision for natural view consistency.
However, these methods commonly encounter challenges in jointly recovering geometry and texture due to the diversity of shape and appearance distributions. Most approaches focus on generating geometry~\cite{nash2020polygen, chen2020bsp} and then proceed through additional stages to obtain appearance~\cite{chen2023text2tex}. 
Recent prevalent approaches involve using 2D supervision, which supervises differentiable renderings from 3D generations using the GT mesh renderings from 3D meshes. For example, ~\cite{muller2023diffrf, ssdnerf} learn NeRF generation from mesh collections like ShapeNet.
However, they require rendering dense multi-views from each mesh, learning respective shape NeRF representation for each training sample from the rendered images, and training the generative diffusion model to model the distribution of NeRF representations for all samples.
Despite achieving reasonable results, we argue that the rendering process is redundant and also introduces information loss. 
In addition, relying on 2D supervision from multi-view renderings also introduce inaccurate reconnstructions, where neural volume rendering is employed to integrate ray colors and calculate rendering loss from pixel colors. We contend that this loss function offers weak supervision, as each pixel color is tasked with supervising the learning of all points' density and color in a ray. 
Particularly when faced with few or imbalanced view observations, this methodology struggles to synthesize an accurate radiance field.

In this work, we aim to synthesize textured shapes while inheriting the advantages of 3D mesh data by denoising the radiance field with guidance from ray color and density values extracted from 3D meshes.
Our method employs individual supervision for each ray point, overseeing both density and color.
To facilitate this, we introduce a module named \textbf{\OURS}, designed to extract theoretically GT sampled ray points from meshes. By utilizing this module to supervise a generative model, such as a modern triplane-based NeRF Diffusion model~\cite{ssdnerf}, our approach exhibits superior performance in both unconditional and conditional 3D textured shape generation tasks. 
In summary, our contributions are:

\begin{enumerate}
\item We propose \OURS, a theoretical derivation to directly create a radiance field from mesh data. Employing \OURS to convert a mesh to a radiance field eliminates the need for rendering multi-view images, avoiding typical imperfections in multi-view reconstruction.
\item We show how Mesh2NeRF can be employed as direct supervision in training generative models with 3D data, especially in applications such as conditional and unconditional NeRF generation.
\end{enumerate}

%% file: AuthorKit/sections_eccv/2relatedworks_eccv.tex
\section{Related Work}

\mypara{NeRF as 3D representation}
Introduced by \cite{mildenhall2021nerf} in 2020, NeRF has become a prominent method for representing scenes. Advances in works like cone tracing with positional encoding \cite{barron2021mip} and proposal multi-layer perceptron (MLP) \cite{barron2022mip} have enhanced neural representation and rendering across various scenarios~\cite{barron2022mip,verbin2022ref, niemeyer2022regnerf, mildenhall2022nerf, mildenhall2022nerf}.
While NeRF has made substantial progress and unlocked numerous applications, there remain significant differences between its representation and the mesh representation widely utilized in traditional computer graphics. Bridging the gap between these two representations is a challenge, and existing methods have explored incorporating meshes into the rendering process to expedite the rendering~\cite{yariv2023bakedsdf, Tang_2023_ICCV}. 
NeRF2Mesh \cite{Tang_2023_ICCV} focuses on reconstructing surface meshes from multi-view images for applications such as scene editing and model composition.
We posit that leveraging existing mesh collections~\cite{chang2015shapenet, collins2022abo, deitke2023objaverse, deitke2024objaverse}, often featuring high-quality data crafted by artists~\cite{polyhaven, skatchfab}, can enhance NeRF/3D applications. 
To achieve this, we propose \OURS{} to directly obtain a radiance field from a textured 3D mesh. By using \OURS{} as direct mesh supervision in NeRF tasks, we aim to harness mesh data to enhance applications utilizing NeRF as the 3D representation.
 
\mypara{NeRF supervision} 
The original NeRF is trained using a pixel loss, where rays sampled from the NeRF are integrated to colors supervised by ground-truth pixel colors. 
However, this approach requires substantial multi-view coverage, as each pixel color is utilized to supervise all 3D points sampled along a ray emitted from the pixel.
To mitigate ambiguity, various methods introduce depth priors as additional supervision to inform networks about scene surfaces~\cite{deng2022depth, roessle2022dense, wei2021nerfingmvs}. In addition to addressing ambiguity, the inference of scene surfaces is often employed for ray termination to prevent further point sampling~\cite{reiser2021kilonerf}.
In contrast, our approach, Mesh2NeRF, accepts textured meshes as input and employs an analytical solution to generate radiance fields. This derived radiance field proves effective in various NeRF tasks, encompassing the fitting of single scenes and the training of NeRF generative models.

\mypara{NeRF Generation with diffusion models}
Recently, NeRF representations have gained popularity in shape generation, especially within diffusion-based methods~\cite{croitoru2023diffusion}. 
In the realm of diffusion models for 3D shape generation, approaches can be broadly categorized into two groups: those extending 2D diffusion priors to the 3D domain~\cite{wang2023score,anciukevivcius2023renderdiffusion,karnewar2023holodiffusion,metzer2023latent,tang2023make}, and those directly applying diffusion processes to 3D data~\cite{shue20233d,muller2023diffrf,cheng2023sdfusion,chou2023diffusion}.
In essence, 2D-lifted diffusion methods inherit the richness of large 2D diffusion models but may grapple with issues of view consistency, while 3D native data diffusion methods exhibit complementary characteristics.
Several diffusion models learn priors from synthetic objects either by learning diffusion weights on pre-calculated NeRF representations~\cite{muller2023diffrf} or by jointly optimizing diffusion weights with each object's NeRF~\cite{ssdnerf}. However, these approaches heavily rely on multi-view observations rendered from meshes.
In contrast, our Mesh2NeRF directly derives analytically correct colors and density values for ray points from meshes. These serve as robust 3D supervision in NeRF generation diffusion models.

%% file: AuthorKit/sections_eccv/3method_eccv.tex
\section{Method}
We provide a brief overview of volume rendering and NeRF in Section~\ref{subsection:nerf}. 
Following that, we delve into how \OURS{} directly derives radiance field representation from textured meshes in Section~\ref{subsec:rf_from_mesh} and elaborate on the application of \OURS{} as direct supervision in NeRFs in Section~\ref{subsec:mesh2nerf_supervision}.
In Section~\ref{subsec:mesh2nerf_gene}, we illustrate the application of \OURS{} in NeRF generation diffusion models.

\begin{figure*}[t]
\centering
\vspace{-0.15in}
\includegraphics[width=1\textwidth]{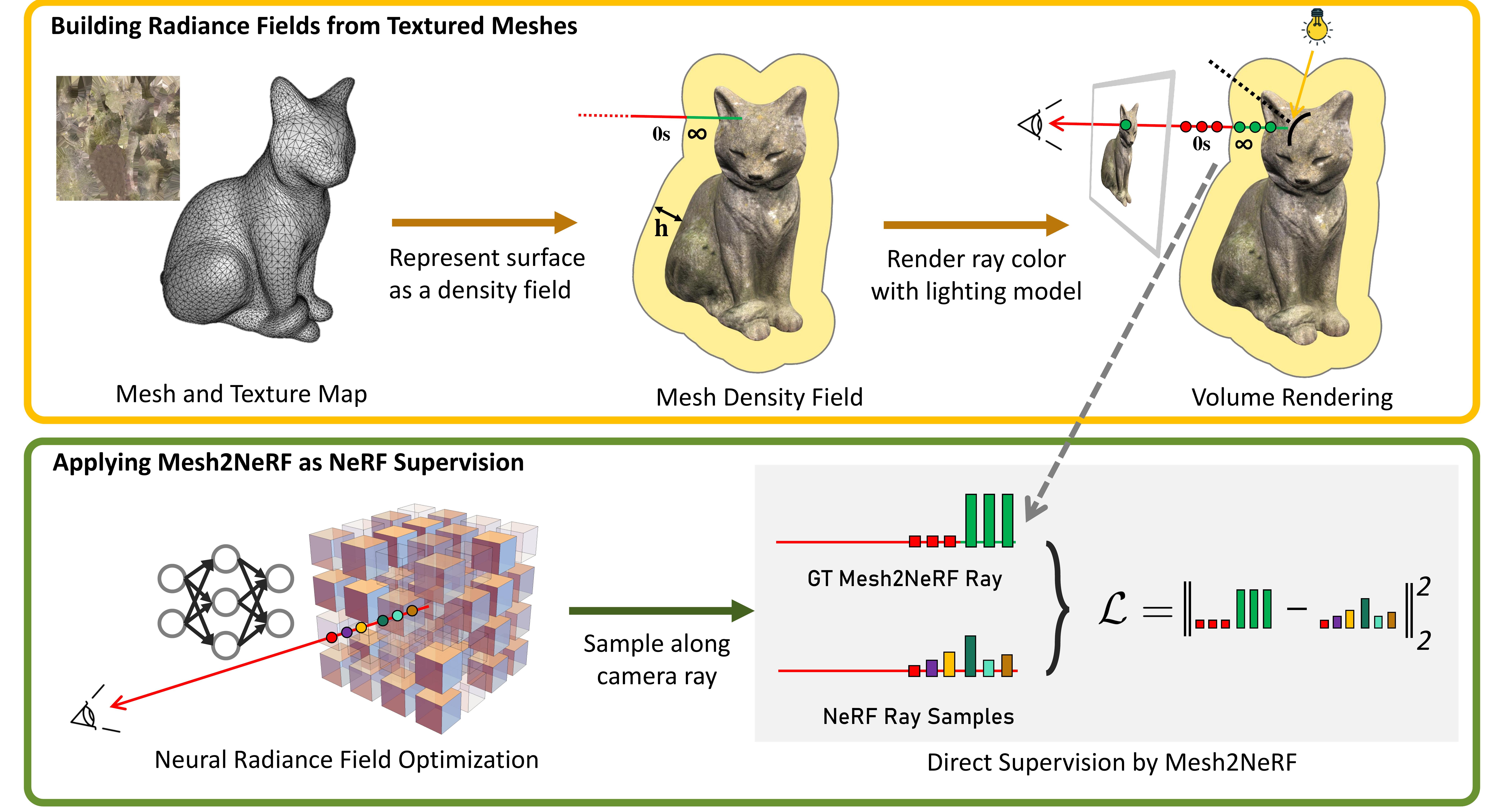}
\vspace{-0.23in}
\caption{
Our method, illustrated above, constructs a ground truth radiance field from a textured mesh. Using a surface-based occupancy function with a distance threshold, we model the scene's density field. View-dependent color is then modeled, considering view direction, surface geometry, and light direction. Integrating samples along the camera ray enables accurate volume rendering from our defined radiance field.
The bottom part showcases \OURS{} as direct 3D supervision for NeRF tasks, where the density and color values of each ray sample supervise NeRF ray samples during optimization.
}
\vspace{-0.2in}
\label{fig:illustration}
\end{figure*}

\subsection{Volume Rendering and NeRFs Revisited}
\label{subsection:nerf}
NeRFs model a 3D scene using a neural network to represent the radiance field. This field takes a 3D position $\mathbf{x}=(x,y,z)$ and viewing direction $\mathbf{d}=(\theta,\phi)$, providing the volume density $\sigma(\mathbf{x})\in\mathbb{R}^+_0$ and RGB color $\mathbf{c}(\mathbf{x},\mathbf{d})$ at that point.
To render the color of a pixel in a target camera view, the 3D position $\mathbf{o}$ of the camera center and the camera viewing direction $\mathbf{d}$ are used to construct the ray $\mathbf{y} = \mathbf{o} + t\mathbf{d}$, with the ray parameter $t\in\mathbb{R}^+_0$. The ray is sampled at $N$ points $\mathbf{y}_i$ with parameter values $t_i$, where the volume density and color are $\sigma_i=\sigma(\mathbf{y}_i)$ and $\mathbf{c}_i=\mathbf{c}(\mathbf{y}_i,\mathbf{d})$. The pixel color is then approximated by the formula from the original NeRF paper~\cite{mildenhall2021nerf,tagliasacchi2022volume}: 
\begin{equation}
    \label{eq:nerf_render}
    \hat{\mathbf{C}}(\mathbf{y}) = \sum_{i=1}^{N} T_i(1-\exp(-\sigma_i\delta_i))\mathbf{c}_i
\end{equation}
Here,
\begin{equation}
   \label{eq:transmittance}
    T_i = \exp\left(-\sum_{j=1}^{i-1}\sigma_j\delta_j\right)
\end{equation}
is the accumulated transmittance along the ray $\mathbf{y}$ to the sample \textit{i} and $\delta_i=t_{i+1}-t_i$ is the distance between adjacent samples.

During NeRF training, the MLP's weights, encoding scene characteristics, are optimized.
The training process aims to optimize MLP weights to match the rendered and GT colors of the rays.
With comprehensive scene coverage in the input images, this straightforward process yields MLP weights that precisely represent the 3D volumetric density and appearance of the scene.
For novel view synthesis, a corresponding image can be rendered from a given camera pose.

\subsection{Radiance Field from Textured Mesh}
\label{subsec:rf_from_mesh}
In the process of constructing a radiance field from a given textured mesh in conjunction with environmental lighting, our objective is to approximate the ideal density field and color field, collectively representing a 3D mesh as a continuous function in $\mathbb{R}^3$. 
In the ideal scenario, the color along each ray is constant and equal to the color of the hitting point on the mesh. The density, on the other hand, resembles a Dirac delta function, implying that only the mesh surface positions have infinite density values, while the density is zero elsewhere.
However, such a density distribution is impractical for training a neural network or for discrete volume rendering with a finite number of samples. 
We therefore adapt our ideal density and
color fields to make them suitable for network training and discrete volume rendering.

\mypara{Density field of mesh surfaces} 
Discrete volume rendering is based on Eq.~\ref{eq:nerf_render}, which assumes the density $\sigma$ and color $\mathbf{c}$ are piecewise constant, as demonstrated in \cite{tagliasacchi2022volume}. We find that the density can be modeled by the top hat functions with a parameter $n$:
\vspace{-2pt}
\begin{equation}
\Delta_n(t)=
\begin{cases}
\frac{n}{2}, & \text{if } |t|<\frac{1}{n} \\
0, & \text{otherwise}
\end{cases}
\end{equation}
These functions provide a density $\sigma(t)$ which is $n\Delta_n(t)$ at every step of the limit process $\lim_{n\rightarrow\infty}$ of a Dirac delta function, and are thus piece-wise constant. 

The height of the density $\sigma(t)$ near the mesh is $n^2/2$ for the top hat functions, which grows indefinitely during the limit process. However, very large numbers are not adequate for a representation with a neural network. Therefore, we use the alpha values
\begin{equation}
\label{eq:alpha}
    \alpha_i = 1-\exp(-\sigma_i\delta_i)
\end{equation}
and reformulate Eqs.~\ref{eq:nerf_render} and \ref{eq:transmittance} to
\begin{align}
\label{eq:color_alpha}
    \hat{C}(\mathbf{y}) = \sum_{i=1}^{N} T_i\alpha_{i}\mathbf{c}_i
\end{align}
with the accumulated transmittance:
\begin{equation}
    T_i =\prod_{j=1}^{i-1}(1-\alpha_j)
\end{equation}
The alpha value $\alpha_i$ denotes how much light is contributed by sample $i$. We refer to the original NeRF paper~\cite{tagliasacchi2022volume} for more details. The alpha value changes from $0$ to $1$ when touching the surface and back to $0$ after passing through it.
Hence, for large $n$, the density $n\Delta_n(t)$ can approximately be represented by:
\begin{equation}
\alpha =
\begin{cases}
    1, & \text{if } d < h \\
    0, & \text{otherwise}
\end{cases} 
\end{equation}
where $\alpha$ is a function of distance to mesh surface $d$; $h$ is the half of the surface thickness.

Our key insight is that we can use occupancy to represent the alpha value $\alpha$. From Eq.~\ref{eq:color_alpha}, we can get the conclusion that $\hat{C}(\mathbf{y}) = \alpha_{i_m}\mathbf{c}_{i_m}$, where $i_m$ is the first sampled point intersecting with the surface, $\alpha_{i_m}=1$ ,and $\mathbf{c}_{i_m}$ is the color of intersection point from the texture mesh.

\begin{figure}[tb!]
\vspace{-0.1in}
\centering
\includegraphics[width=0.999\textwidth]{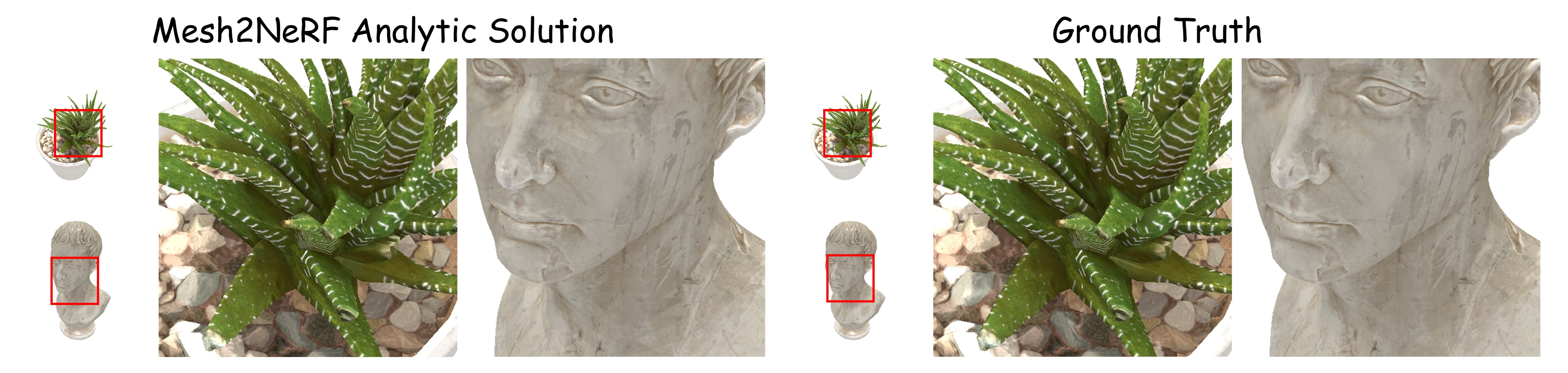}
\vspace{-0.2in}
\caption{Volume renderings from the defined radiance fields of Mesh2NeRF analytic solution. Our rendering results are very close to the ground truth, indicating that our defined radiance fields can serve as an effective ground truth representation for NeRFs.}
\vspace{-0.1in}
\label{fig:denfied_render}
\end{figure}

\mypara{Modeling the color field} In \OURS, we represent the color of sampled points along each ray as a constant using a BRDF model, e.g., we use the Phong model in our experiments while any BRDF can be applied. For any sampled point on the ray with direction $\mathbf{v}$, its color is defined as the color $\mathbf{c_i}$ of the first hitting point of the ray and mesh surface.
In Fig.~\ref{fig:denfied_render}, we qualitatively compare volume rendering outcomes between our defined \OURS  fields (obtained through \OURS analytic solution) with ground truth renderings. The latter corresponds to the mesh rendering conducted under identical lighting and shading conditions.
Our method generates accurate, high-quality view-dependent renderings, showcasing the precision of our discretely defined density and color distribution along the ray.

\subsection{\OURS~as Supervision in NeRFs}
\label{subsec:mesh2nerf_supervision}
Our approach enables the generation of density and color attributes for each sampled point along the rays,  allowing direct supervision of these values in the outputs of NeRF MLP. For predicted density  $\hat{\mathcal{\sigma}}$ and color $\hat{\mathbf{c}}_{i}$ along the ray, alpha $\hat{\mathcal{\alpha}}_{i}$ can be computed using Eq.~\ref{eq:alpha}.
The overall optimization loss term is composed of alpha and color terms:
\begin{equation}
    \mathcal{L}_\mathit{alpha} = \sum_{i=1}^{N} \lvert \hat{\alpha}_{i} - \alpha_{i} \rvert^{2}
\end{equation}
\begin{equation}
    \mathcal{L}_\mathit{color} = \sum_{i=1}^{N} \lVert \hat{\mathbf{c}}_{i} - \mathbf{c}_{i} \rVert^{2}_{2}
\end{equation} 
Optionally, the overall optimization loss can include a term, $\mathcal{L}_\mathit{integral}$, accounting for the difference in the ray color integral:
\begin{equation}
\label{eq:integral_loss}
    \mathcal{L}_\mathit{integral} = \left\lVert  \sum_{i=1}^{N} \hat{\alpha}_{i}\hat{\mathbf{c}}_i\prod_{j=1}^{i-1}(1\!-\!\hat{\alpha}_j)
    \!-\!
    \sum_{i=1}^{N} \alpha_{i}\mathbf{c}_i\prod_{j=1}^{i-1}(1\!-\!\alpha_j)
    \right\rVert^{2}_{2}
\end{equation}
The final optimization loss term is given by:
\begin{equation}
\label{eq:mesh2nerf_loss}
    \mathcal{L} = \mathcal{L}_\mathit{alpha} + w_\mathit{color}\mathcal{L}_\mathit{color} + w_\mathit{integral}\mathcal{L}_\mathit{integral}
\end{equation}
where $w_\mathit{color}$ and $w_\mathit{integral}$ are weighting terms.

\mypara{Space sampling for neural radiance field optimization}
\OURS offers robust supervision for a neural radiance field generated from a mesh. 
As depicted in the lower part of Fig.~\ref{fig:illustration}, points are sampled along a ray and fed to an MLP to predict density and color (like NeRFs). 
We emulate a virtual camera in unit spherical to define ray origins and directions  similar to NeRFs but use ray casting to compute intersections with the geometry.
For efficient sampling of the fields, we implement queries on top of the bounding volume hierarchy acceleration structures of the Embree library \cite{embree}. 
This information optimizes the sampling of points inside the volume. 
For rays intersecting the surface, we implement stratified sampling of points along the ray in both the empty scene space and within a narrow band at a distance $h$ to the surface. 
If a ray does not intersect the mesh, we randomly sample points in the domain along the ray.
\OURS directly supervises the MLPs at each sampled point using the density and color fields generated from the mesh through the loss terms $\mathcal{L}_\mathit{alpha}$ and $\mathcal{L}_\mathit{color}$.

\subsection{\OURS in NeRF Generation Tasks}
\label{subsec:mesh2nerf_gene}
To facilitate 3D generation tasks,  we advocate for the incorporation of \OURS{} into the NeRF generation framework. We build on our generative framework based on the single-stage diffusion NeRF model~\cite{ssdnerf}, which combines a triplane NeRF auto-decoder with a triplane latent diffusion model. 
The pivotal innovation lies in replacing the rendering loss, which relies on multi-view observations rendered from meshes, with our direct supervision.

\mypara{SSDNeRF Revisited}
In the context of multiple scene observations, generalizable multi-scene NeRFs can be trained by optimizing per-scene codes and shared parameters, as proposed in~\cite{chen2023factor}. 
This optimization is achieved by minimizing the rendering loss $\mathcal{L}_\mathit{rend}$ associated with each observation image.
Consequently, the model is trained as an auto-decoder\cite{park2019deepsdf}, where each scene code is interpreted as a latent code.
To enhance the expressiveness of latent representations, a latent diffusion model (LDM) is employed to learn the prior distribution in the latent space. The LDM introduces Gaussian perturbations into the code at the diffusion time step. Subsequently, a denoising network with trainable weights is utilized to remove noise during the diffusion stage, predicting a denoised code. The LDM is trained using a simplified L2 denoising loss $\mathcal{L}_\mathit{diff}$.
In the SSDNeRF framework, the training objective aims to minimize the variational upper bound on the negative log-likelihood (NLL) of observed data:
\begin{equation}
    \mathcal{L}_\mathit{ssdnerf} = w_\mathit{rend}\mathcal{L}_\mathit{rend} + w_\mathit{diff} \mathcal{L}_\mathit{diff}
\end{equation}
Here, the scene codes, prior parameters, and NeRF decoder parameters are jointly optimized, with $w_\mathit{rend}$ and $w_\mathit{diff}$ serving as weighting factors for the rendering and diffusion losses, respectively.

With trained diffusion prior, a variety of solvers (such as DDIM~\cite{song2020denoising}) can be used to recursively denoise a random Gaussian noise, until reaching the denoised state in the unconditional sampling process. When the generation is conditioned, e.g., from single-view or sparse-view observations, the sampling process is guided by the gradients of rendering loss from known observation. For more details about the image-guided sampling and fine-tuning, we refer to SSDNeRF~\cite{ssdnerf}.

\mypara{Supervision with \OURS} When mesh data is directly used, one can replace the rendering loss $\mathcal{L}_\mathit{rend}$ to \OURS loss as Eq.~\ref{eq:mesh2nerf_loss} during the training process. In the conditional generation stage, since the surface distribution is not available from observation conditions, we use the rendering loss $\mathcal{L}_\mathit{rend}$ to compare standard volume rendering with the observation pixel to guide the generation.

%% file: AuthorKit/sections_eccv/4results_eccv.tex
\section{Results}
We illustrate that our direct supervision of radiance fields from meshes significantly improves performance in NeRF optimization and generation tasks.
We showcase results of a single scene fitting using Mesh2NeRF, comparing it with NeRF pipelines in Section~\ref{subsec:single_scene_fitting}. Importantly, we then demonstrate the utility of Mesh2NeRF in diverse 3D generation tasks in Section~\ref{subsec:cond_generation} and Section~\ref{subsec:uncond_generation}.

\begin{figure*}[tb!]
\centering
\includegraphics[width=0.99\textwidth]{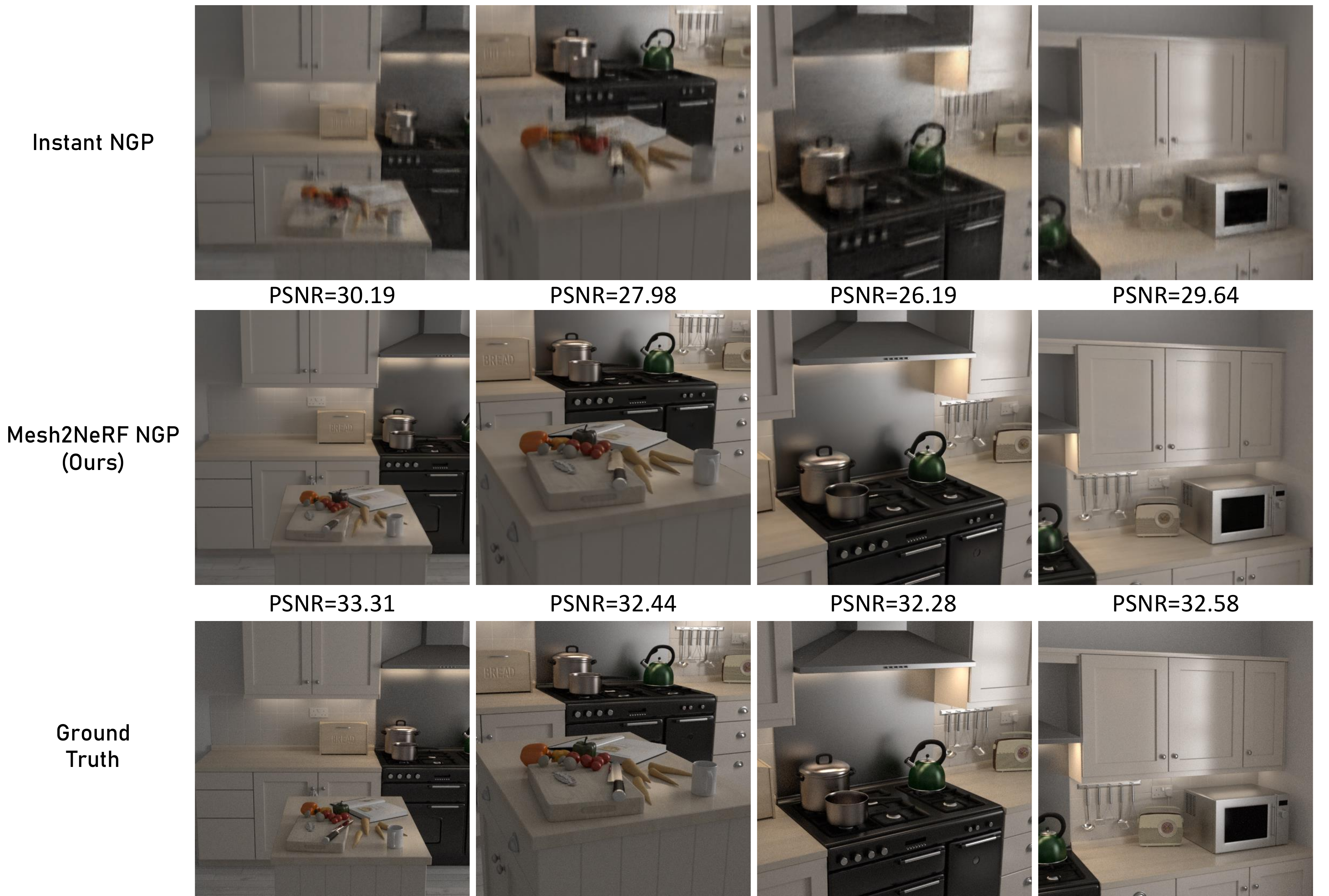}
\caption{Comparison of single scene fitting on the \textit{Country-Kitchen} scene.
Our results showcase higher accuracy and a superior ability to capture finer details in renderings when compared to the baseline method (Mesh2NeRF~NGP vs Instant~NGP).
}
\label{fig:vis_blend}
\end{figure*}

\subsection{Single Scene Fitting}
\label{subsec:single_scene_fitting}
This section compares \OURS with traditional NeRF methods in representing a single scene via a neural network. \OURS directly supervises the neural network from mesh data, while traditional NeRFs use rendering technologies and then fuses information from renderings to the radiance field.

We evaluate our method on scenes from Amazon Berkeley Objects (ABO) \cite{collins2022abo}, Poly Haven Models~\cite{polyhaven}, Sketchfab Scenes~\cite{skatchfab} and the \textit{Country-Kitchen} scene~\cite{country-kitchen}. 
The ABO dataset contains twelve realistic 3D models of household objects. 
Each object is rendered at 512$\times$512 pixels from viewpoints sampled on a sphere, covering the object's surface area (90 views as input and 72 for testing). 
We also show results on six hyper-real models from the Poly Haven library, which consists of quality content designed by artists. Each scene is rendered at 512$\times$512 with 90 input and 120 test views.
We use the \textit{Country-Kitchen} scene to evaluate complex indoor scene under realistic lighting and material effect. It is also rendered at 512$\times$512 pixels with 1,208 input views and 205 test views.
For \OURS, we use the textured mesh from both sets by setting a fixed point light. 
We evaluate our approach with three different encodings and parametric data structures for storing trainable feature embeddings, \ie, the network part of NeRFF~\cite{mildenhall2021nerf}, TensoRF~\cite{chen2022tensorf}, and Instant NGP~\cite{muller2022instant} with our \OURS direct supervision using the corresponding mesh. We denote these three methods by Mesh2NeRF NeRF, Mesh2NeRF TensoRF, and Mesh2NeRF NGP, respectively.

In Table~\ref{tab:single_object_results}, we report mean PSNR/SSIM (higher is better) and LPIPS (lower is better) for objects in both the ABO and Poly Haven datasets. 
Our method consistently outperforms prior work on both datasets. 
For the ABO dataset, all three variants of our method surpass baseline optimization across all evaluation metrics. 
Mesh2NeRF NGP, employing Instant NGP style multi-resolution hash table encodings, achieves the best performance. 
Compared to the raw Instant NGP version, Mesh2NeRF NGP exhibits notable improvements ($+3.12$ PSNR, $+0.041$ SSIM, and $-0.021$ LPIPS). 
Similarly, on the Poly Haven dataset, Mesh2NeRF NGP delivers the best results.

We present qualitative results in Fig.~\ref{fig:vis_blend}, illustrating the superiority of our method over NeRF baseline (Instant NGP) in representing single scenes on challenging \textit{Country-Kitchen}.
Mesh2NeRF NGP exhibiting fewer artifacts and providing more precise renderings.
For additional implementation details, results and analysis, please refer to the supplemental material.

\begin{table}[tb]
\centering
\caption{View synthesis results for fitting a single scene on ABO and Poly Haven datasets. Our comparison involves NeRF methods and \OURS, utilizing identical network architectures while differing in supervision style.
}
\scalebox{0.82}{
\begin{tabular}{lcccccc}
\Xhline{2\arrayrulewidth}
\multirow{2}{*}{Method}& \multicolumn{3}{c}{\textit{ABO Dataset}}   & \multicolumn{3}{c}{\textit{Poly Haven Dataset}}\\
\cmidrule(lr){2-4}
\cmidrule(lr){5-7}
& \makebox[0.1\textwidth][c]{PSNR \textuparrow} & \makebox[0.1\textwidth][c]{SSIM \textuparrow} & 
\makebox[0.1\textwidth][c]{LPIPS \textdownarrow} & \makebox[0.1\textwidth][c]{PSNR \textuparrow} & \makebox[0.1\textwidth][c]{SSIM \textuparrow} & \makebox[0.1\textwidth][c]{LPIPS \textdownarrow} \\
\hline
NeRF & 25.09  & 0.882 & 0.137 & 21.24 & 0.689 & 0.422  \\
TensoRF  & 31.33& 0.944  & 0.032  & 23.32 & 0.748 & 0.260   \\
Instant NGP  & 30.16 & 0.928 & 0.039  & 24.39 & 0.779 & 0.185  \\
\hline
\OURS NeRF  & 32.40
& 0.942
& 0.044
& 22.75 & 0.710 & 0.296 \\
\OURS TensoRF  & 32.00
& 0.957  
& 0.024  
& 23.97 & 0.761 & 0.220 \\
\OURS NGP  & 
\textbf{33.28} 
& \textbf{0.969}
& \textbf{0.018}
&  \textbf{25.30}  
& \textbf{0.825}
& \textbf{0.129}
\\
\Xhline{2\arrayrulewidth}
\end{tabular}
}
\label{tab:single_object_results}
\end{table}

\subsection{Conditional Generation}
\label{subsec:cond_generation}
This section presents experimental results of NeRF generation conditioned on images of unseen objects from the ShapeNet Cars and Chairs~\cite{chang2015shapenet}, as well as the KITTI Cars~\cite{geiger2012we}.
These datasets pose unique challenges, with the ShapeNet Cars set featuring distinct textures, the ShapeNet Chair set exhibiting diverse shapes, and KITTI Cars representing real-world inference data with a substantial domain gap when compared to the training data of the generative model.

\mypara{Implementation information}
We adopt SSDNeRF~\cite{ssdnerf} as the framework for sparse-view NeRF reconstruction tasks since it achieves state-of-the-art NeRF generation performance.
In its single-stage training, SSDNeRF jointly learns triplane features of individual scenes, a shared NeRF decoder, and a triplane diffusion prior. 
The optimization involves a pixel loss from known views. 
For \textit{Ours}, we replace the pixel loss with 
Mesh2NeRF supervision (Eq.~\ref{eq:mesh2nerf_loss}) from the train set meshes. 
We generate training data for ShapeNet Cars and Chairs using the same lighting and mesh processes as in our Mesh2NeRF setup. 
SSDNeRF is trained using the official implementation; our model is trained with the same viewpoints as in the SSDNeRF experiments, without additional ray sampling, ensuring a fair comparison.
Our evaluation primarily centers on assessing the quality of novel view synthesis based on previously unseen images.
PSNR, SSIM, and LPIPS are evaluated to measure the image quality. 

\begin{figure*}[t!]
\centering
\includegraphics[width=0.90\textwidth]{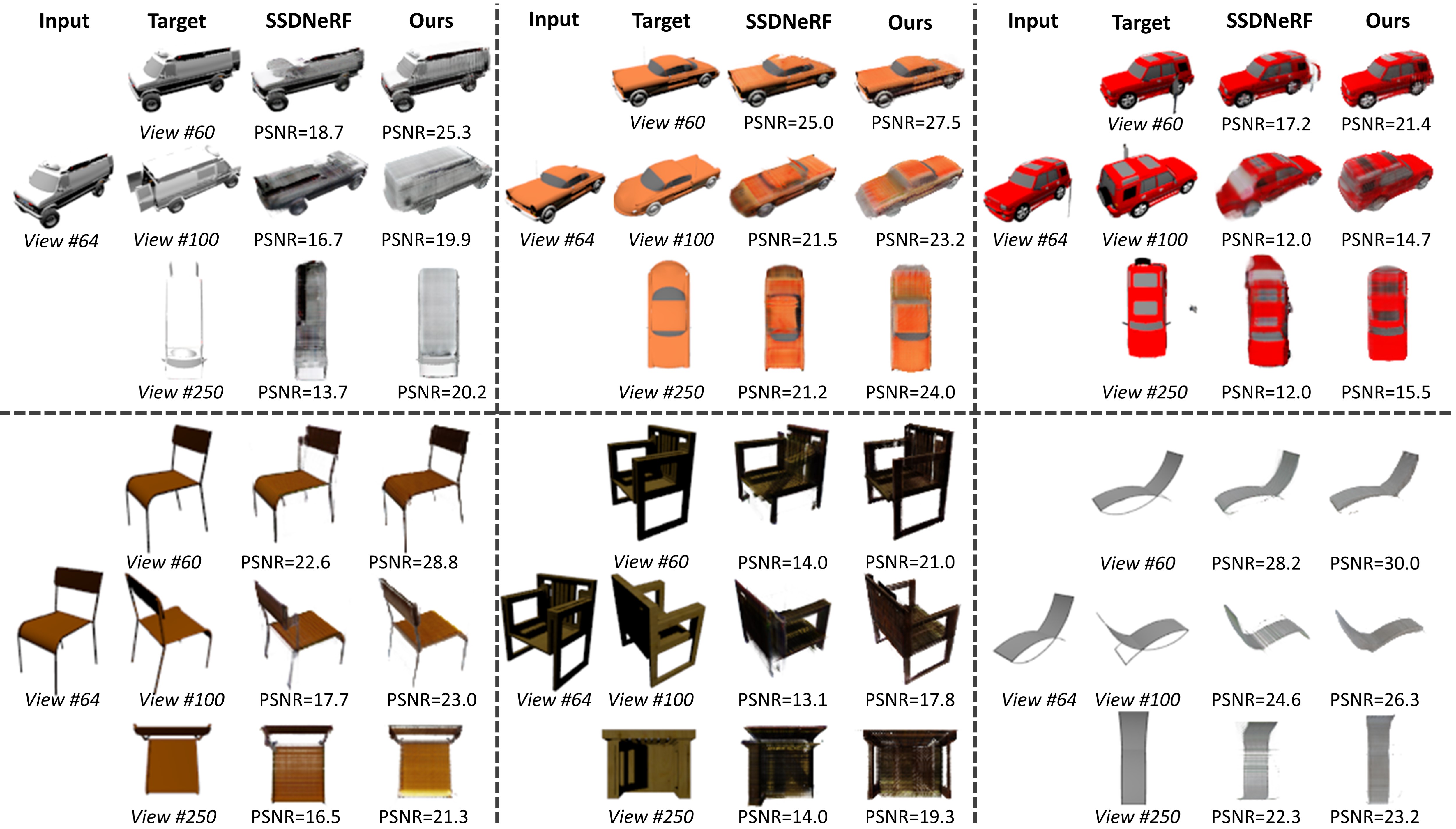}
\caption{Qualitative comparison of NeRF generation conditioned on single-view for unseen objects in ShapeNet Cars and Chairs between SSDNeRF and our method.  
Our approach enables more accurate novel views.
}
\label{fig:ssdnerf_single}
\end{figure*}

\begin{figure*}[t]
\centering
\includegraphics[width=0.90\textwidth]{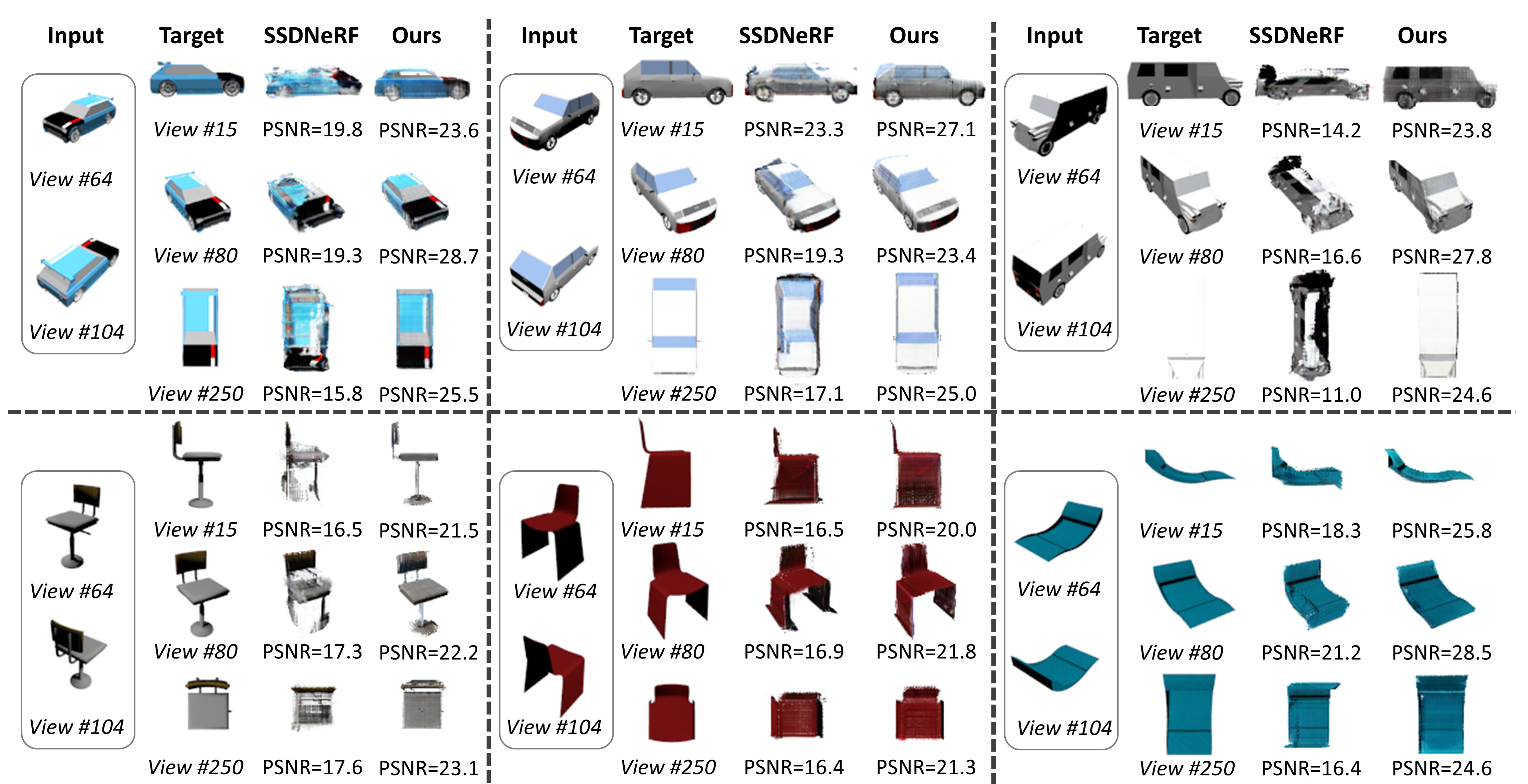}
\caption{Qualitatively comparison of NeRF generation conditioned on two-view for unseen objects in ShapeNet Cars and Chairs. Three novel views are displayed for each two-view input. Our approach enables more accurate novel views.}
\label{fig:2view_condition}
\end{figure*}

\begin{table}[tb!]
\caption{Conditional NeRF generation results on ShapeNet Cars and Chairs. 
} 
\begin{center}
\scalebox{0.7}{
\setlength{\tabcolsep}{0.60em}
\begin{tabular}{lcccccccccccc}
\toprule
& \multicolumn{3}{c}{Cars 1-view} & \multicolumn{3}{c}{Cars 2-view} & \multicolumn{3}{c}{Cars 3-view} & \multicolumn{3}{c}{Cars 4-view} \\
\cmidrule(lr){2-4}
\cmidrule(lr){5-7}
\cmidrule(lr){8-10}
\cmidrule(lr){11-13}
& \makebox[0.07\textwidth]{PSNR \textuparrow} & \makebox[0.07\textwidth]{SSIM \textuparrow} & \makebox[0.07\textwidth]{LPIPS \textdownarrow} & \makebox[0.07\textwidth]{PSNR \textuparrow} & \makebox[0.07\textwidth]{SSIM \textuparrow} & \makebox[0.07\textwidth]{LPIPS \textdownarrow} & \makebox[0.07\textwidth]{PSNR \textuparrow} & \makebox[0.07\textwidth]{SSIM \textuparrow} & \makebox[0.07\textwidth]{LPIPS \textdownarrow} & \makebox[0.07\textwidth]{PSNR \textuparrow} & \makebox[0.07\textwidth]{SSIM \textuparrow} & \makebox[0.07\textwidth]{LPIPS \textdownarrow} \\
\midrule
SSDNeRF & 21.09 & 0.881 & 0.104 & 24.67 &  0.926 & \textbf{0.071}  & 25.71 & 0.934 & 0.069 & \textbf{26.54} & 0.939 & 0.067 \\
Ours & \textbf{21.78} & \textbf{0.893} & \textbf{0.101} & \textbf{24.98} & \textbf{0.932} & 0.072 & \textbf{25.89} & \textbf{0.942} & \textbf{0.066} & 26.51 & \textbf{0.945} & \textbf{0.062} \\
\midrule
 & \multicolumn{3}{c}{Chairs 1-view} & \multicolumn{3}{c}{Chairs 2-view} & \multicolumn{3}{c}{Chairs 3-view} & \multicolumn{3}{c}{Chairs 4-view}\\
 \cmidrule(lr){2-4}
\cmidrule(lr){5-7}
\cmidrule(lr){8-10}
\cmidrule(lr){11-13}
& \makebox[0pt]{PSNR \textuparrow} & \makebox[0pt]{SSIM \textuparrow} & \makebox[0pt]{LPIPS \textdownarrow}  & \makebox[0pt]{PSNR \textuparrow} & \makebox[0pt]{SSIM \textuparrow} & \makebox[0pt]{LPIPS \textdownarrow}  & \makebox[0pt]{PSNR \textuparrow} & \makebox[0pt]{SSIM \textuparrow} & \makebox[0pt]{LPIPS \textdownarrow}  & \makebox[0pt]{PSNR \textuparrow} & \makebox[0pt]{SSIM \textuparrow} & \makebox[0pt]{LPIPS \textdownarrow}\\
\midrule
SSDNeRF & 19.05 &  0.853 & 0.133 & 19.65 & 0.859 & 0.136 & 20.42 & 0.863 & 0.149 & 21.84 & 0.884 & 0.134 \\
Ours & \textbf{19.62} & \textbf{0.859} & \textbf{0.128} & \textbf{22.22} & \textbf{0.888} & \textbf{0.112} & \textbf{23.03} & \textbf{0.900} & \textbf{0.116} & \textbf{22.68} & \textbf{0.907} & \textbf{0.109}\\
\bottomrule
\end{tabular}
}
\end{center}
\label{tab:spare-view}
\end{table}

\mypara{Comparative evaluation} 
In Table~\ref{tab:spare-view}, a comprehensive comparison between our method and the state-of-the-art SSDNeRF is provided. Our method demonstrates overall superiority in conditional NeRF reconstruction tasks when provided with sparse-view conditions (ranging from one to four views) for both the Car and Chair categories.
Fig.~\ref{fig:ssdnerf_single} visually illustrates the impact of \OURS supervision on the accuracy of NeRFs in the context of single-view condition generation.
Notably, our method outperforms SSDNeRF, especially when dealing with objects that possess unconventional forms (e.g., the first car with a roof of the same color as the background, and the third red car exhibiting input noise). SSDNeRF struggles in such scenarios, yielding unreasonable geometries.
Qualitative comparisons in the setting of 2-view conditions are presented in both Fig.~\ref{fig:2view_condition}. In these scenarios, our approach consistently outshines SSDNeRF, producing superior results across various instances.

\begin{figure*}[t!]
\centering
\includegraphics[width=0.90\textwidth]{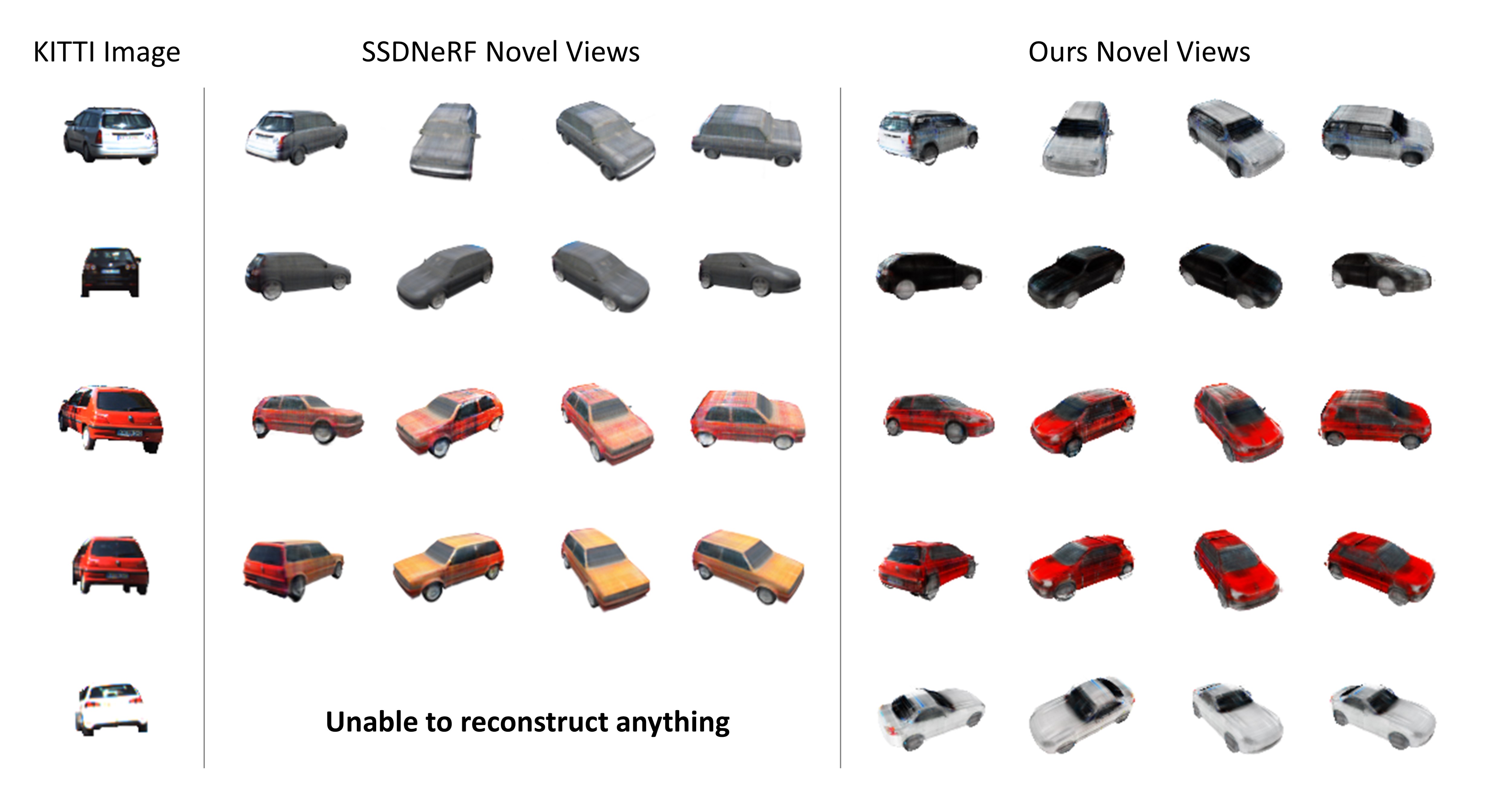}
\caption{Qualitative comparison of novel view synthesis on in-the-wild car images from the KITTI dataset. Four novel views are displayed for each single-view input. Our approach enables more reasonable novel views.
}
\label{fig:kitti}
\end{figure*}

\mypara{Single-view NeRF generation from real images} 
We also conduct a comparison with SSDNeRF using real KITTI car data for conditional generation based on a single-view input. 
This task poses a challenge because the generative model is trained on synthetic ShapeNet cars, resulting in a significant domain gap. 
The input images are obtained from the KITTI 3D object detection dataset~\cite{geiger2012we} using annotated 3D bounding boxes. To align them with the ShapeNet Cars dataset, we utilize provided bounding box dimensions and poses. Segmentation masks remove the background~\cite{heylen2021monocinis}. Images are cropped and resized to $128\times128$.
The processed image conditions our generative model, guiding the generation of a radiance field for rendering novel views around the cars.
In Fig.~\ref{fig:kitti}, we show qualitative examples of novel view synthesis.
Both SSDNeRF and our model successfully reconstruct the car's shape in most cases.
However, SSDNeRF fails in the last row sample due to color similarity between the car and the background. Our model excels in reconstructing the radiance field of the car in this challenging scenario. 
Regarding the generated car NeRFs, our approach produces more accurate textures with improved global consistency and better correspondence to the input image.
In contrast, SSDNeRF tends to generate textures only in the vicinity of the observation area, leaving other regions textureless or with incorrect colors. 
Our results exhibit greater realism and coherence across the entire scene, showcasing the superior generalization capability of our model, especially in the face of substantial domain gaps.

\subsection{Unconditional Radiance Field Synthesis}
\label{subsec:uncond_generation}
We conduct evaluation for unconditional generation using the Objaverse Mugs collection~\cite{deitke2023objaverse}, which consists of 153 mug models. 
The Mugs dataset presents a challenge in generating shape geometry and realistic textures due to the limited training samples. 
The baseline SSDNeRF is trained on rendered images from 50 views of each mesh, and \OURS is trained on meshes with the same viewpoints.
Both models underwent training for 100,000 iterations.

\begin{figure*}[t!]
\centering
\includegraphics[width=0.90\textwidth]{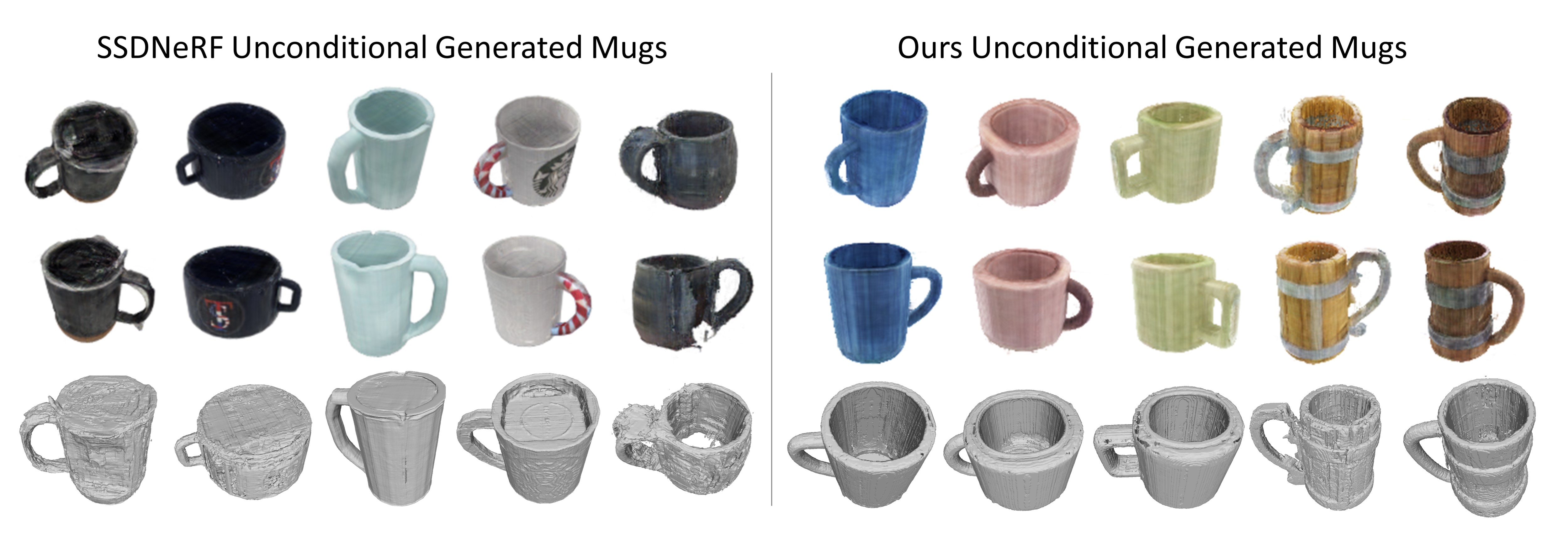}
\caption{Qualitative comparison between unconditional generative models training on Objaverse Mugs. We present two renderings and extracted mesh from each generated NeRF. Our approach exhibits a notable advantage in reasonable geometry, particularly for shapes that suffer from self-occlusions when trained only with image supervision.
}
\label{fig:uncond_mugs}
\end{figure*}

As shown in Fig.~\ref{fig:uncond_mugs}, both SSDNeRF and \OURS generate reasonable NeRFs in the rendered images.
However, when extracting meshes from NeRFs of SSDNeRF, inaccuracies are evident; the geometries of its unconditional generation do not faithfully represent the real geometry of the training samples or real data. 
Notably, traditional NeRF supervision struggles to enable the generative model to capture precise geometric details, such as whether a cup is sealed.
In contrast, \OURS synthesizes diverse and realistic geometries corresponding to the rendered images. 
This underscores the robustness of \OURS supervision, proving instrumental in generating physically plausible 3D content.

\subsection{Limitations}
While \OURS effectively employs direct NeRF supervision for NeRF representation and generation tasks, several limitations persist. 
Similar to NeRF, which bakes lighting information into appearance, we model the result of the Phong lighting
model in the generated radiance field for compatibility with existing approaches.
Another limitation stems from the ray sampling employed by baselines due to the dependence on rendered images. Hence, future work should delve into more efficient sampling techniques that leverage ground truth geometry and appearance, bypassing the need for sampling schemes relying on virtual cameras. 
Furthermore, acknowledging the diversity of mesh data, the application of Mesh2NeRF to generative models can span across various categories of mesh data, contributing to the development of a more universal and robust prior.

%% file: AuthorKit/sections_eccv/5conclusion_eccv.tex
\section{Conclusion}
We have introduced \OURS{}, a new approach for directly converting textured mesh into radiance fields.
By leveraging surface-based density and view-dependent color modeling, our method ensures faithful representation of complex scenes.
Employing \OURS{} as direct 3D supervision for NeRF optimization yields superior performance, enhancing accuracy and detail in view synthesis across diverse hyper-real scenes.
Importantly, the versatility of our method extends to various generalizable NeRF tasks, encompassing both conditional and unconditional NeRF generation. 
When combined with available mesh data, \OURS{} mitigates traditional NeRF supervision shortcomings, leading to enhanced overall performance on synthetic datasets and real-world applications.
\OURS{} is applicable to various NeRF representations, such as frequency encoding (NeRF), hash encoding (NGP), tensorial (TensoRF), and triplane (SSDNeRF), consistently improving respective tasks. 
We believe that this work contributes valuable insights to the integration of mesh and NeRF representations, paving the way for new possibilities in 3D content generation.

%% file: AuthorKit/sections_eccv/6appendix_eccv.tex
\appendix
\section{Implementation Details}
\subsection{Details of Single Scene Fitting}

\mypara{Datasets} 
For the ABO dataset, we use twelve objects representing various household categories\footnote{\scriptsize \url{https://amazon-berkeley-objects.s3.amazonaws.com/index.html}}, including a chair (ABO item ID: B075X4N3JH), a table (B072ZMHBKQ), a lamp (B07B4W2GY1), a vase (B07B8NZQ68), a planter (B075HWK9M3), a dumbbell set (B0727Q5F94), a light fixture (B07MBFDQG8), a cabinet (B008RLJR2G), a basket set (B07HSMVFKY), a sofa (B073G6GTK7), and two beds (B075QMHYV8 and B07B4SCB4H).
Each object is normalized to [-1,1] in three dimensions.
For baselines that require rendered images and camera poses to train, we render each object from cameras distributed on a sphere with a radius of 2.7 using PyTorch3D~\cite{johnson2020accelerating}. We use the lighting defined by PyTorch3D PointLights, with a location in [0, 1, 0], and the ambient component [0.8, 0.8, 0.8], diffuse component [0.3, 0.3, 0.3], and specular component [0.2, 0.2, 0.2].
In our method, rendered RGB images are not used. Instead, we directly utilize the same normalized meshes and the light information.

For the Poly Haven dataset, we use six realistic models\footnote{\scriptsize \url{https://polyhaven.com/models}}, including \href{https://polyhaven.com/a/potted_plant_01}{a potted plant}, \href{https://polyhaven.com/a/BarberShopChair_01}{a barber chair}, \href{https://polyhaven.com/a/CoffeeCart_01}{a coffee chart}, \href{https://polyhaven.com/a/chess_set}{a chess set}, \href{https://polyhaven.com/a/concrete_cat_statue}{a concrete cat statue}, and \href{https://polyhaven.com/a/garden_hose_wall_mounted_01}{a garden hose}. 
Rendered images for baseline NeRF training inputs are normalized to the [-1,1] cube. We render each object from cameras distributed on a sphere with a radius of 2.0. The point light is located at [0, 2, 0], with ambient, diffuse, and specular components set to [1.0, 1.0, 1.0], [0.3, 0.3, 0.3], and [0.2, 0.2, 0.2], respectively. Mesh2NeRF training employs the same normalized meshes and lighting.
For the SketchFab scene, we use \href{https://sketchfab.com/3d-models/entree-du-chateau-des-bois-francs-8dc01ff203bc4f02ac74c0b00f441be5}{Entrée du château des Bois Francs} and \href{https://sketchfab.com/3d-models/chateau-des-bois-francs-822e334e36824a1095f67a778cf5b214}{Château des Bois Francs}
. The rendering settings remain consistent with Poly Haven data, with a camera distance of 1.5.

\mypara{Training}
We implement our method in Python using the PyTorch framework. We leverage PyTorch3D \cite{johnson2020accelerating} and Open3D \cite{zhou2018open3d} for processing mesh data and performing essential computations in Mesh2NeRF.
Training on both the ABO and Poly Haven datasets comprises 50,000 iterations with a batch size of 1024 rays. 
We employ the Adam optimizer \cite{kingma2014adam} with a 1cycle learning rate policy \cite{smith2019super}, where the max learning rate is set to $1\times 10^{-3}$, the percentage of the cycle by 0.001, and the total number of steps is 50,000.
For each sampled point, we assign a weight $w_{color}$ of 1 if the sample lies within the surface thickness; otherwise, $w_{color}$ is set to 0. Additionally, we fix $w_{integral}$ at 10 during the training process.

\mypara{Sampling}
During training, we employ ray casting to compute the intersection distance for each ray and the input mesh.
For each ray, we conduct stratified sampling, obtaining 512 points distributed along the ray. 
In the case of rays intersecting the surface, an additional 512 points are sampled within the surface thickness using a random sampling approach. For rays that do not intersect the surface, we randomly sample another set of 512 points along the ray.
For each sampled point, Mesh2NeRF can extract ground truth density (or alpha value) and color information directly from the input mesh. This procedure ensures the availability of accurate supervision data for training.
In the inference phase, given a specific view, we utilize the camera's intrinsic parameters to determine the origin and direction of the ray. Subsequently, we uniformly sample points along the ray within the object cube. In a scenario with a surface thickness of 0.005, we sample 800 points along each target ray.

\subsection{Details of Conditional Generation}

\mypara{Data setting}
In the main manuscript, we evaluate NeRF conditional generation on ShapeNet Cars and Chairs, and additionally perform inference on KITTI Cars.
For ShapeNet Cars and Chairs, the model is trained ontheir respective training sets.
During inference, for the 1-view setup, we use view 64 as the NeRF input and evaluate the other 249 views out of 250.
For the 2-view setup, we use views 64 and 104 as the NeRF input and evaluate the other 248 views.
We additionally evaluate the results of 3-view and 4-view setups in Section~\ref{subsec:appendix_sparse_view}.
For the 3-view setup, we use views 0, 125, and 250 as the NeRF input and evaluate the other 247 views.
For the 4-view setup, we use views 0, 83, 167, and 250 as the NeRF input and evaluate the other 246 views.

\mypara{Training and evaluation}
For SSDNeRF results, we adhere closely to its official implementation\footnote{\scriptsize \url{https://github.com/Lakonik/SSDNeRF}} during model training, ensuring consistency with our training data. 
The evaluation employs the same test set as SRN \cite{sitzmann2019scene}.
For Mesh2NeRF, Blender is utilized to obtain color and intersection distance information for each viewpoint. Additional details are available in the ShapeNet rendering repository\footnote{\scriptsize \url{https://github.com/yinyunie/depth_renderer}}. 
In Mesh2NeRF supervision during training, we use a surface thickness of 0.01. Along each ray, we employ a stratified sampling strategy to distribute 32 points along the ray, along with an additional random sampling of 32 points within the surface thickness. Both $w_\mathit{color}$ and $w_\mathit{integral}$ are set to 1. In the inference stage, we evenly sample 400 points along each target ray for volume rendering.

\mypara{Traning and inference time}
We train our generative model using two RTX A6000 GPUs, each processing a batch of 8 scenes. On average, 80K training iterations for ShapeNet experiments take around 40 hours, and 10K training iterations for Objaverse mugs take around 5 hours. 
For inference, under the unconditional generation setting using 50 DDIM steps, sampling a batch of 8 scenes takes 4 sec on a single RTX A6000 GPU.
And under the conditional generation setting, and 130 sec for reconstructing a batch of 8 scenes from a single-view condition.

\section{Additional Results}

\begin{figure}[tb]
\centering
\includegraphics[width=0.7\textwidth]{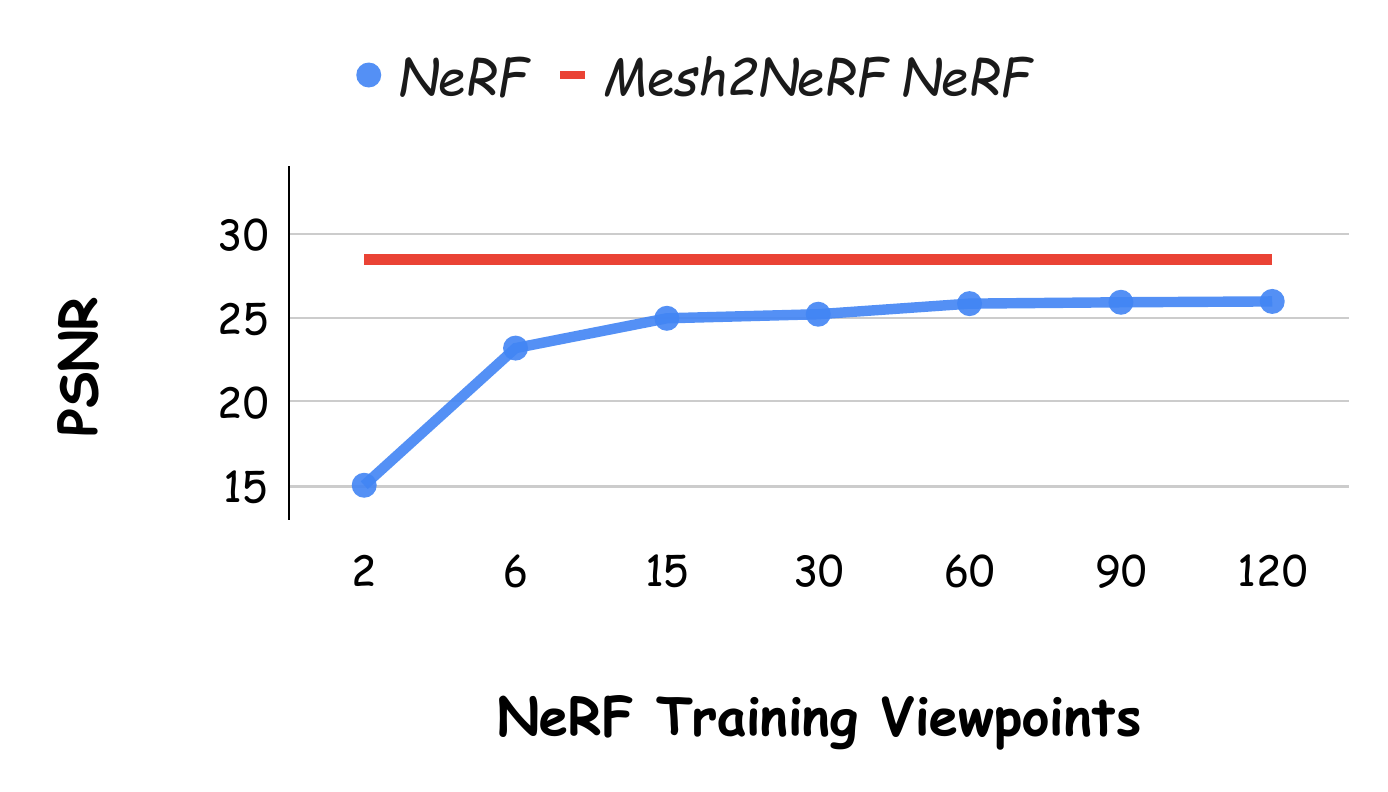}
\caption{Comparing \OURS with NeRF trained with the different viewpoint numbers. The performance of NeRF saturates as the number of viewpoints increases, \ie, adding viewpoints results in diminishing returns at some point. Overall, the performance is worse than \OURS, which does not require renderings and is aware of the whole scene content.}
\label{fig:nerf_views}
\end{figure}

\begin{table*}[tb!]
\centering
\caption{Ablation of integral loss $L_\mathit{integral}$ of Mesh2NeRF NGP on the Poly Haven dataset.}
\scalebox{0.7}{
\begin{tabular}{c|c|cccccc|c}
\toprule
& Variant & \makebox[0.16\textwidth][c]{Potted Plant} & \makebox[0.16\textwidth][c]{Barber Chair} & \makebox[0.16\textwidth][c]{Coffee Chart} & \makebox[0.16\textwidth][c]{Chess Set} & \makebox[0.16\textwidth][c]{Cat Statue} & \makebox[0.16\textwidth][c]{Garden Hose} & \textit{Average} \\
\midrule
\multirow{5}{*}{PSNR \textuparrow} 
& V1 & 11.53 & 27.42 & 11.90 & 7.02 & 12.42 & 7.09 & 12.90 \\
& V2 & 19.26 & 23.43 & 21.74 & 22.62 & 23.91 & 20.59 & 21.93 \\
& V3 & \textbf{23.96} & 25.76 & 24.93 & 25.32 & 24.53 & 22.68 & 24.53 \\
& V4 & 23.66 & \textbf{26.79} & \textbf{26.96} & \textbf{26.00} & \textbf{24.64} & \textbf{23.77} & \textbf{25.30}\\
& V5 & 13.67 & 7.08 & 11.91 & 25.95 & 14.84 & 8.73 & 13.70 \\
\midrule
\multirow{5}{*}{SSIM \textuparrow}
& V1 & 0.657 & \textbf{0.926} & 0.739 & 0.563 & 0.485 & 0.423 & 0.632 \\
& V2 & 0.744 & 0.892 & 0.849 & 0.851 & 0.676 & 0.731 & 0.791 \\
& V3 & 0.819 & 0.900 & 0.881 & 0.873 & \textbf{0.689} & 0.750 & 0.819 \\
& V4 & \textbf{0.829} & 0.907 & \textbf{0.901} & \textbf{0.877} & 0.683 & \textbf{0.756} & \textbf{0.825} \\
& V5 & 0.642 & 0.535 & 0.739 & 0.884 & 0.556 & 0.471 & 0.638 \\
\midrule
\multirow{5}{*}{LPIPS \textdownarrow} 
& V1 & 0.448 & 0.096 & 0.446 & 0.596 & 0.618 & 0.620 & 0.477 \\
& V2 & 0.123 & \textbf{0.078} & 0.110 & \textbf{0.076} & \textbf{0.236} & \textbf{0.106} & \textbf{0.122} \\
& V3 & 0.088 & 0.100 & 0.108 & \textbf{0.076} & 0.261 & 0.114 & 0.125 \\
& V4 & \textbf{0.087} & 0.101 & \textbf{0.105} & 0.084 & 0.268 & 0.128 & 0.129 \\
& V5 & 0.353 & 0.548 & 0.446 & 0.087 & 0.459 & 0.514 & 0.401 \\
\bottomrule
\end{tabular}
}
\label{tab:integral_ablation}
\end{table*}

\begin{table}[tb]
\centering
\caption{Mesh2NeRF NeRF results vary with defined surface thickness on the ABO dataset. Smaller thickness yields better MLP optimization performance, with 0.0050 and 0.0025 providing comparable results. We choose 0.0050 as the default thickness for subsequent experiments.}
\scalebox{0.95}{
\begin{tabular}{cccc}
\toprule
\makebox[0.2\textwidth][c]{Surface Thickness} & \makebox[0.2\textwidth][c]{PSNR \textuparrow} & \makebox[0.2\textwidth][c]{SSIM \textuparrow} & \makebox[0.2\textwidth][c]{LPIPS \textdownarrow}\\
\midrule
0.0200 & 27.18 & 0.919  & 0.058\\
0.0100  & 27.18 & 0.929 & 0.053\\
0.0050  & 28.40 & 0.933 & \textbf{0.049}\\
0.0025  & \textbf{28.93} & \textbf{0.934} & 0.056\\
\bottomrule
\end{tabular}
}
\label{tab:ablation_thickness}
\end{table}

\subsection{More Results on Single Scene Fitting}

\mypara{Impact of integral loss}
In this ablation study, we investigate the impact of the integral loss during optimizing neural radiance fields in Mesh2NeRF.  
We compare five variants of our method, denoted as (V1) through (V5), except for (V1), we maintain $w_\mathit{alpha}=1$ and $w_\mathit{color}=1$. (V1): only uses the integral loss $\mathcal{L}_\mathit{integral}$, without the alpha loss $\mathcal{L}_\mathit{alpha}$ and the color loss the integral loss $\mathcal{L}_\mathit{color}$;
(V2): excludes the integral loss $\mathcal{L}_\mathit{integral}$; (V3) with the integral loss and uses a weight of $w_\mathit{integral}=1$; (V4) with the integral loss and uses a weight of $w_\mathit{integral}=10$. (V5) with the integral loss and uses a weight of $w_\mathit{integral}=100$.
We evaluate the PSNR, SSIM, and LPIPS of each sample test view and provide overall average results across all test views in the dataset.
As shown in Table~\ref{tab:integral_ablation}, (V4) achieves the highest average PSNR and SSIM, while (V2) exhibits slightly superior performance in terms of LPIPS. Consequently, we select V4 as the default configuration for our Mesh2NeRF.

\mypara{Comparing using NeRFs with different numbers of viewpoints}
We compare \OURS~NeRF with NeRF trained using varying numbers of views on the chair object from the ABO dataset.
In Fig.~\ref{fig:nerf_views}, we chart the mean PSNR for NeRF w.r.t. training views rendered from the mesh, and \OURS, directly optimized from mesh. 
The evaluation results of NeRF exhibit improvement from a few viewpoints to dozens of views (\eg, covering most of the surface), converging with increasing views. \OURS captures more comprehensive object information even compared to NeRF at convergence, acting as an upper bound.

\mypara{Ablation study on surface thickness} 
In Table~\ref{tab:ablation_thickness}, we self-compare surface thickness as a hyperparameter defining mesh resolution.
For objects normalized to [-1,1] in three dimensions, we vary the surface thickness from 0.02 to 0.0025.
Results show average PSNR, SSIM, and LPIPS for test views at $256\times 256$ resolution. A smaller thickness captures finer details but demands denser sampling for rendering. 
Consequently, we set 0.005 as the default surface thickness for our method.

\mypara{Lighting model} We apply Phong model in our experiments, while \OURS can accommodate any BRDF and lighting.
The only adjustment required is to change the function computing the color value of the ray hit point. For instance, if Spherical Harmonics lighting replaces the point lighting, the analytic solution (as illustrated in Fig.~3) can also achieve an SSIM over 0.99.

\mypara{Baseline with geometry information} 
We compare NeRF baseline with and without GT depth supervision during fitting on the Sketchfab scene \textit{Entrée du château des Bois Francs}.
Results (Table~\ref{tab:baseline_depth}) show improvements in Instant NGP with depth supervision in both dense (same as Sec.~4.1) and sparse (30 views) settings. More improvement from depth is gained in the limited views setup (+0.44 PSNR vs. +0.21). Even with depth, Instant NGP is still worse than Mesh2NeRF NGP.
\begin{table}[tb]
\centering
\caption{Comparison with Instant NGP with additional depth supervision. Ours (Mesh2NeRF NGP) outperforms Instant NGP in both limited and dense view settings.}
\scalebox{0.9}{
\begin{tabular}{ccl}
\toprule
\makebox[0.4\textwidth][c]{Method} & \makebox[0.25\textwidth][c]{Depth loss} &\makebox[0.25\textwidth][l]{PSNR} \\
\midrule
\multirow{2}{*}{Instant NGP (limited views)} & No               & 15.75\\
 & Yes & 16.19 \textit{\textcolor{red}{(+0.44)}}                      \\
\midrule
\multirow{2}{*}{Instant NGP (dense views)} & No & 19.51 \\
 & Yes & 19.72 \textit{\textcolor{red}{(+0.21)}} \\
\midrule
Mesh2NeRF NGP & -  &  20.42 \\
\bottomrule
\end{tabular}
}
\label{tab:baseline_depth}
\end{table}

\begin{figure*}[tb!]
\centering
\includegraphics[width=1\textwidth]{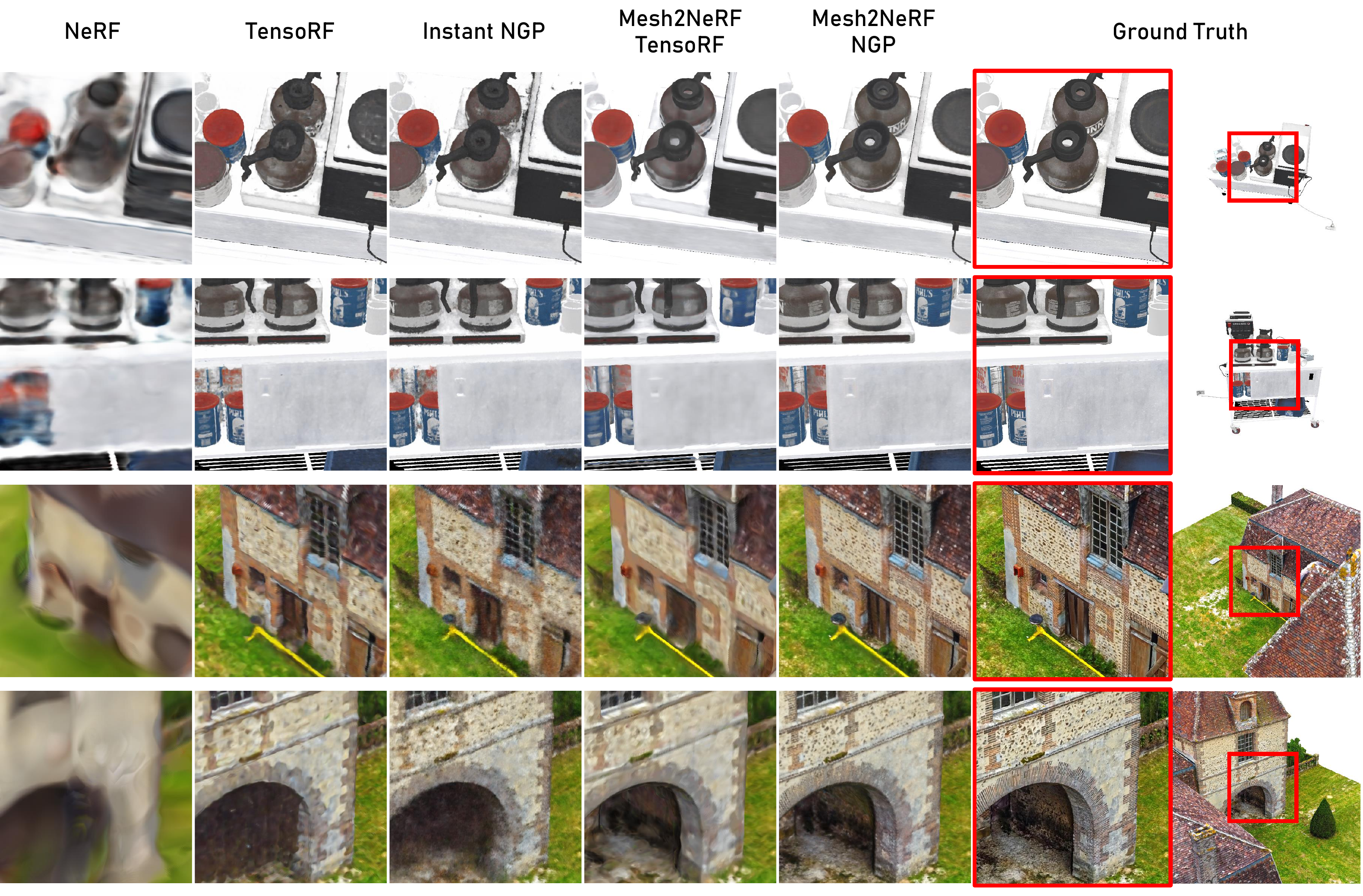}
\caption{Comparison of single scene fitting for scenes from Poly Haven and Sketchfab.  
For each visualized scene, we present two renderings of test views by each method.
Our results showcase higher accuracy and a superior ability to capture finer details in renderings when compared to the baseline methods (Mesh2NeRF~TensoRF vs. TensoRF and Mesh2NeRF~NGP vs. Instant~NGP).
}
\label{fig:vis_vsnerf}
\end{figure*}

\begin{figure*}[th]
\centering
\scalebox{1}{
\includegraphics[width=1\textwidth]{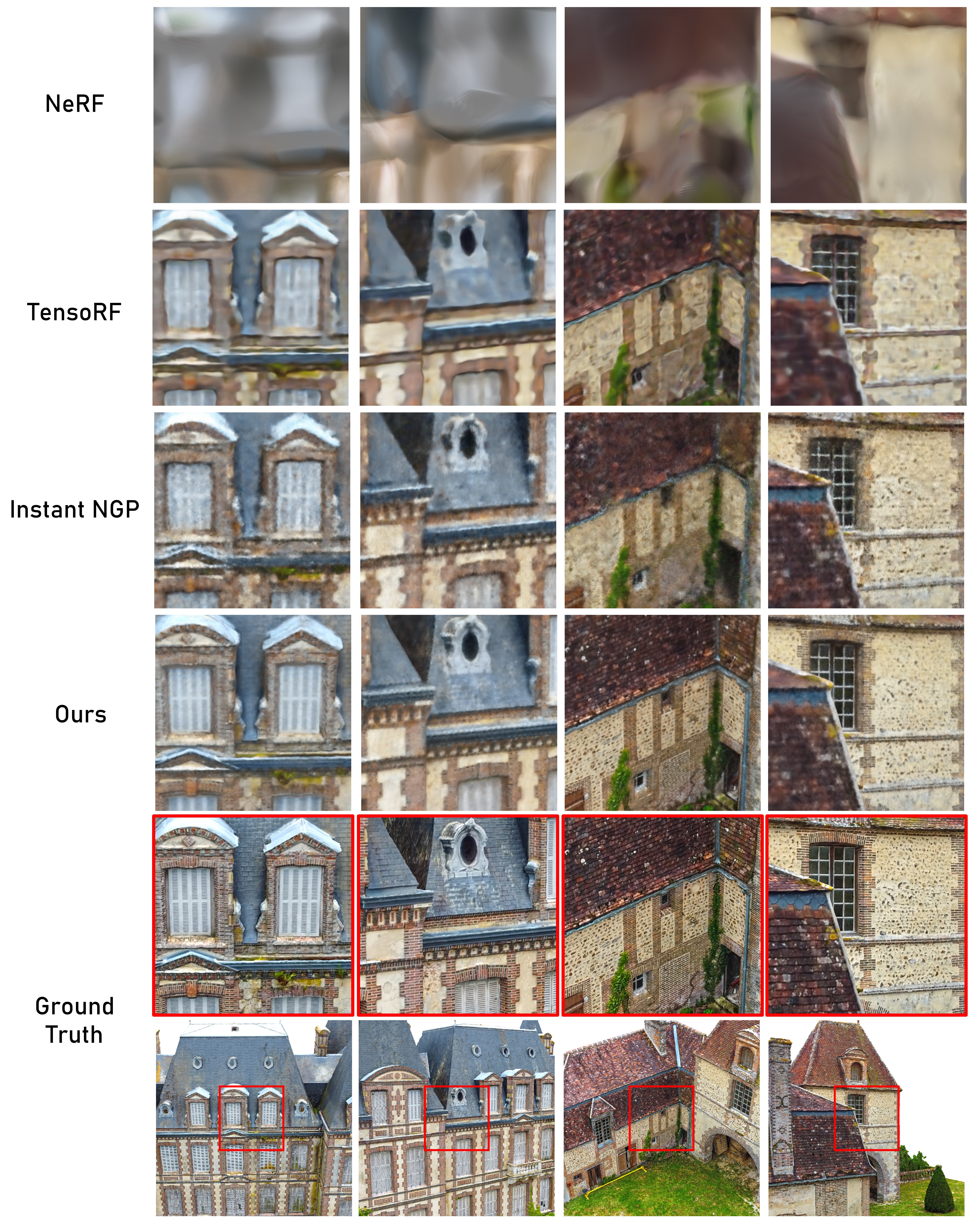}
}
\caption{Comparison on test views for scenes from the Sketchfab dataset. Ours (Mesh2NeRF NGP) outperforms NeRF baseliens in the displayed challenging scenes.}
\label{fig:sf_supp_visual}
\end{figure*}

\begin{figure*}[th]
\centering
\scalebox{1.0}{
\includegraphics[width=0.9\textwidth]{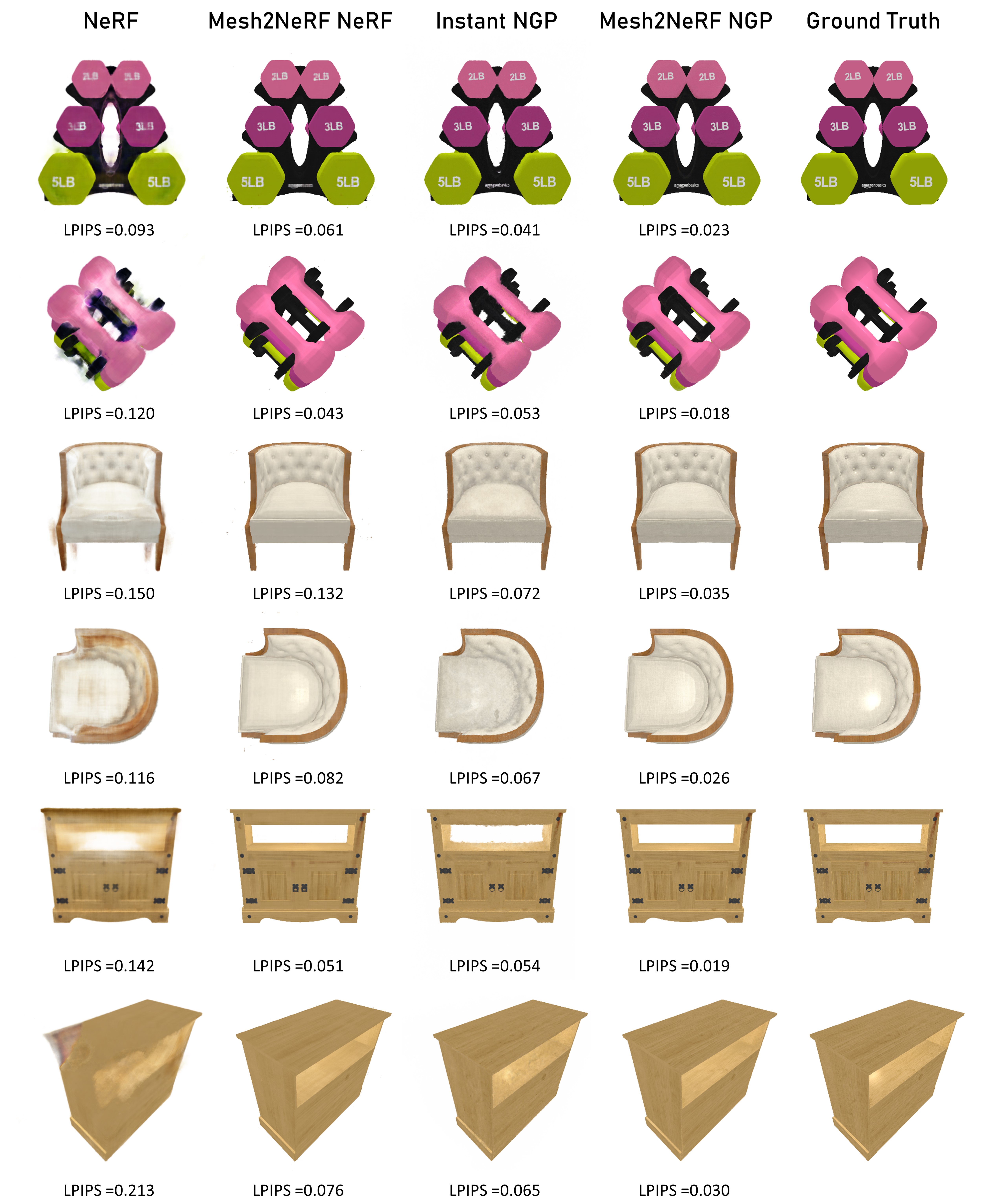}
}
\caption{Comparison on test views for scenes from ABO dataset. 
For every visualized object, we show two renderings from each
method. Our results (Mesh2NeRF NeRF vs. NeRF and Mesh2NeRF NGP vs. Instant NGP) are more accurate and capture finer details in renderings compared to the baselines using the same network architecture.}
\label{fig:abo_visual}
\end{figure*}

\mypara{More qualitative comparsions} 
We present qualitative results in Fig.~\ref{fig:vis_vsnerf}, illustrating the superiority of our method over NeRF baselines in representing single scenes on challenging \textit{Coffee Cart} from Poly Haven and \textit{Entrée du château des Bois Francs} from Sketchfab.
Mesh2NeRF TensoRF supasses the original TensoRF, indicating that \OURS supervision contributes to the neural network's ability to learn accurate scene representation.
We present qualitative results from the Sketchfab dataset in Fig.\ref{fig:sf_supp_visual}. Renderings from two views for each object are compared with NeRF baselines, accompanied by corresponding PSNR values.
Similarly, Fig.\ref{fig:abo_visual} showcases qualitative results on the ABO dataset, with renderings from two views for each object compared alongside corresponding LPIPS values. These figures demonstrate that Mesh2NeRF NeRF outperforms NeRF baselines.

\begin{figure*}[th]
\centering
\scalebox{1}{
\includegraphics[width=1\textwidth]{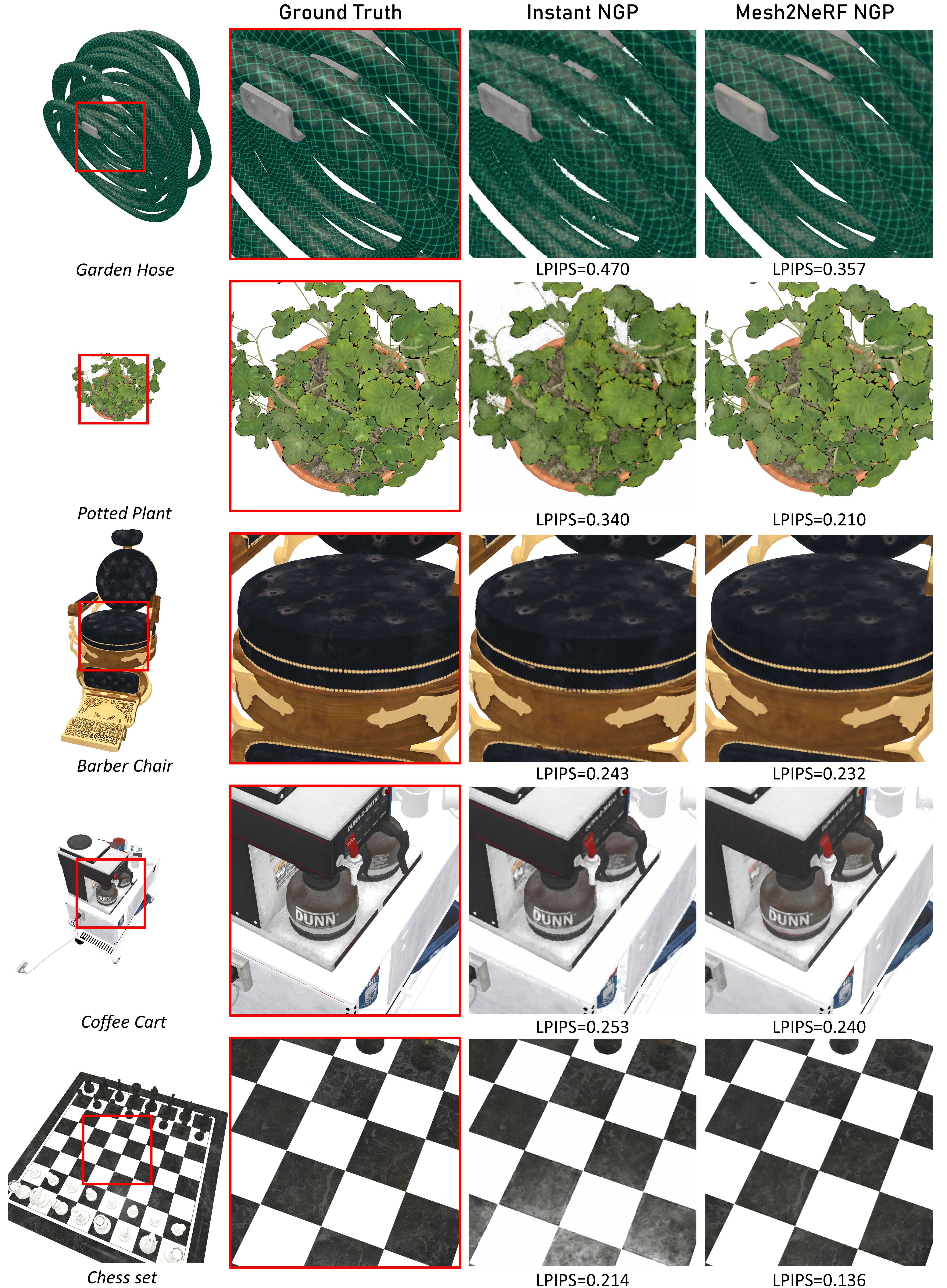}
}
\caption{Comparison on test views for scenes from the Poly Haven dataset at a resolution of $1024\times1024$. Without modifying the training process, Mesh2NeRF NGP outperforms Instant NGP in generating high-resolution renderings.}
\label{fig:ph_supp_visual}
\end{figure*}

\mypara{High-resolution volume rendering}
Our method is not constrained by resolution during rendering, as evidenced by our evaluation setting. In Fig.~\ref{fig:ph_supp_visual}, we compare \OURS{} NGP and Instant NGP renderings at resolutions of $1024\times1024$. Our approach consistently outperforms Instant NGP, demonstrating its ability to achieve high-resolution results during inference.

\subsection{More Results on Conditional Generation}
\label{subsec:appendix_sparse_view}

\begin{figure*}[th]
\centering
\includegraphics[width=1\textwidth]{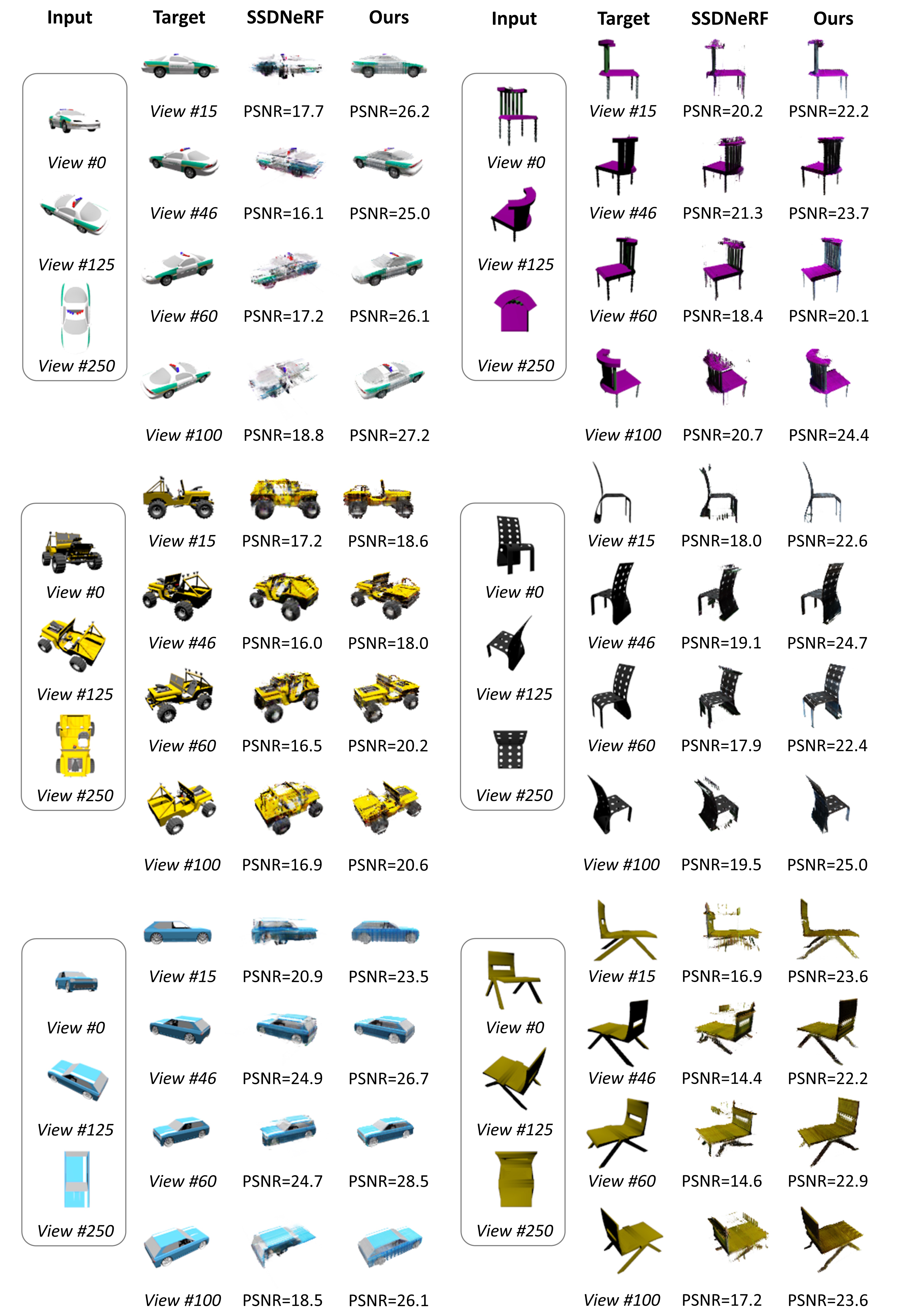}
\caption{Qualitatively comparison of NeRF generation conditioned on 3-view input on unseen objects in ShapeNet Cars (left part of the figure) and Chairs (right part).}
\label{fig:3view}
\end{figure*}

\begin{figure*}[th]
\centering
\includegraphics[width=0.98\textwidth]{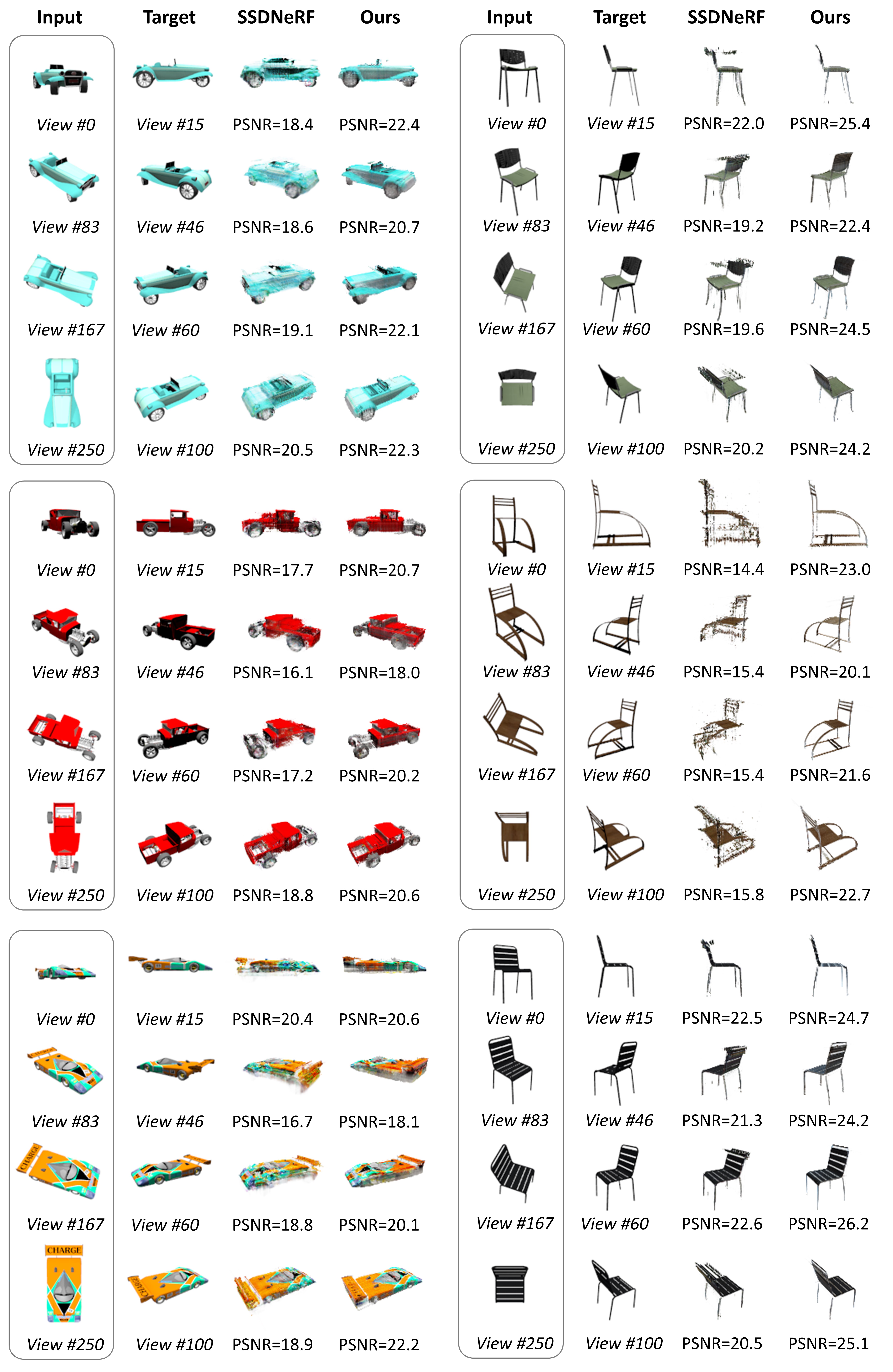}
\caption{Qualitatively comparison of NeRF generation conditioned on 4-view input on unseen objects in ShapeNet Cars (left part of the figure) and Chairs (right part).}
\label{fig:4view}
\end{figure*}

\begin{figure*}[th!]
\centering
\includegraphics[width=1\textwidth]{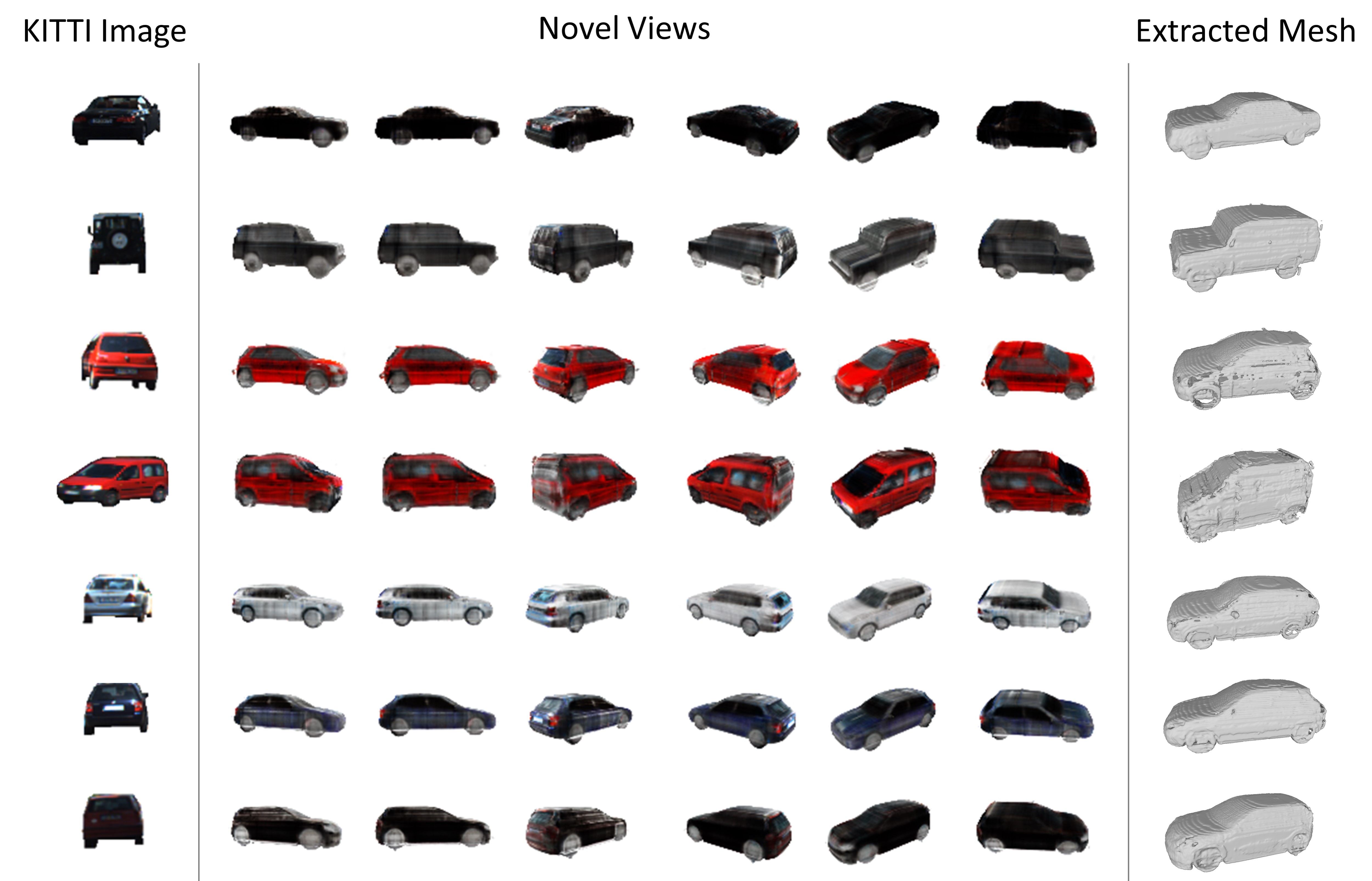}
\caption{Qualitatively comparing conditional NeRF generation of KITTI Cars images. We show the input in-the-wild image, rendered novel views of the generated NeRFs, and extracted meshes.}
\label{fig:KITTI_SUPP}
\end{figure*}

\mypara{More spare-view NeRF conditional generation results}
We present qualitative comparisons of NeRF generation conditioned on sparse-view images on previously unseen objects in ShapeNet Cars and Chairs between SSDNeRF and our method as the supervision.
In Figure~\ref{fig:3view}, we showcase the results of NeRF generation conditioned on 3-view inputs in both ShapeNet Cars and Chairs.
For a broader perspective, Figure~\ref{fig:4view} illustrates the results of NeRF reconstructions from 4-view inputs in both ShapeNet Cars and Chairs.
In Figure~\ref{fig:KITTI_SUPP}, we provide addition NeRF generation results conditioned on single-view KITTI Cars real images, utilizing the model trained on the synthetic ShapeNet Cars. This highlights the generalization capability of \OURS supervision in NeRF generation tasks, even when faced with significant domain gaps.

\begin{figure*}[th!]
\centering
\includegraphics[width=0.99\textwidth]{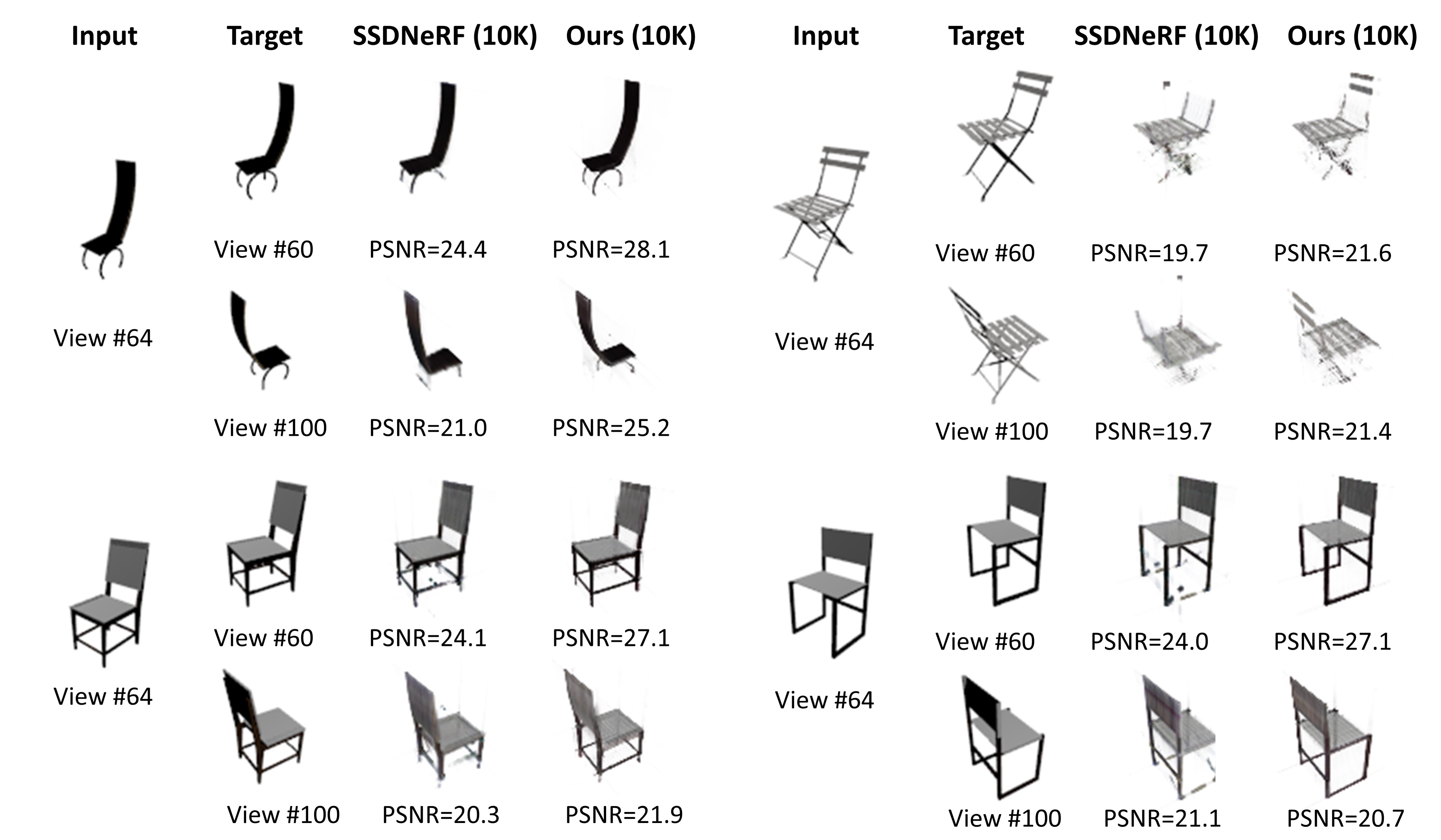}
\caption{Qualitatively comparing single-view NeRF reconstruction of ShapeNet test Chair images at 10,000 training iterations. \OURS{} outperforms SSDNeRF, highlighting the effectiveness of our direct supervision.}
\label{fig:ssdnerf_10k}
\end{figure*}

\mypara{Study on early-stage results} 
Our method offers direct supervision to the 3D radiance field, facilitating faster model convergence during training. 
To illustrate this, we compare the single-view reconstruction results for chair samples at 10,000 iterations (out of a total of 80,000 iterations). 
As shown in Figure~\ref{fig:ssdnerf_10k}, our reconstructed novel views demonstrate improved accuracy and reduced floating noise.

